\documentclass[11pt,3p,review,authoryear]{elsarticle}

\usepackage{microtype}
\usepackage{graphicx}
\usepackage{subfig}
\usepackage{booktabs}
\usepackage{appendix}
\usepackage{multicol}	
\usepackage{multirow} 
\usepackage{hyperref}
\usepackage{amsmath}
\usepackage{amssymb}
\usepackage{amsthm}
\usepackage{nccmath}
\usepackage{mathtools}
\usepackage{subfig}
\usepackage{adjustbox}

\begin{document}

\begin{frontmatter}

\title{The cost of ensembling: is it always worth combining?}

\author[a]{Marco Zanotti\corref{cor}} 
\ead{zanottimarco17@gmail.com}
\address[a]{Department of Economics, Management and Statistics, University of Milano-Bicocca, Milan, Italy}
\cortext[cor]{Corresponding author}

\begin{abstract}
    Given the continuous increase in dataset sizes and the complexity of forecasting models, 
    the trade-off between forecast accuracy and computational cost is emerging as an extremely 
    relevant topic, especially in the context of ensemble learning for time series forecasting. 
    To assess it, we evaluated ten base models and eight ensemble configurations across two 
    large-scale retail datasets, considering both point and probabilistic accuracy 
    under varying retraining frequencies. We showed that ensembles consistently improve 
    forecasting performance, particularly in probabilistic settings. However, these gains come 
    at a substantial computational cost, especially for larger, accuracy-driven ensembles. We 
    found that reducing retraining frequency significantly lowers costs, with minimal impact 
    on accuracy, particularly for point forecasts. Moreover, efficiency-driven ensembles offer 
    a strong balance, achieving competitive accuracy with considerably lower costs compared 
    to accuracy-optimized combinations. Most importantly, small ensembles of two or three models 
    are often sufficient to achieve near-optimal results. These findings provide practical 
    guidelines for deploying scalable and cost-efficient forecasting systems, supporting the 
    broader goals of sustainable AI in forecasting. Overall, this work shows that careful 
    ensemble design and retraining strategy selection can yield accurate, robust, and 
    cost-effective forecasts suitable for real-world applications.
\end{abstract}


\end{frontmatter}


\section{Introduction} \label{sec:intro}

Ensemble learning has become a powerful and widely adopted approach in time 
series forecasting due to its ability to improve predictive accuracy and 
robustness by aggregating the outputs of multiple models. The fundamental 
idea behind ensemble methods is that combining forecasts from diverse models 
can help balance their individual strengths and weaknesses, thereby reducing 
the risk of overfitting and mitigating model-specific biases or errors. This 
approach leverages the principle that while individual models might make 
different types of errors, their combination can average out these errors. 
One of the most intriguing and empirically validated findings in the literature 
is that simple ensemble strategies, such as taking the mean or median of 
forecasts, often outperform more sophisticated combination methods like 
stacking (also known as meta-learning) or weighted ensembles. This phenomenon, 
known as the forecast combination puzzle \citep{forecombpuzzle}, suggests 
that in many cases, the added complexity of estimating optimal weights or 
training a meta-learner does not yield meaningful gains in accuracy and can 
even lead to overfitting, especially when the number of forecast points 
is limited. Indeed, the success of simple combinations is partly due to their 
ability to remain robust in the presence of model misspecification, parameter 
uncertainty, and changing data dynamics, issues that more complex ensemble 
methods may struggle to accommodate effectively.

In general, the benefits of ensemble learning are numerous: it improves the forecast 
accuracy, enhances the model stability across time and datasets, and increases 
resilience to structural breaks or shifts in the data. Ensembles also provide 
a practical hedge against model uncertainty, allowing practitioners to rely less 
on the assumption that any single model is correctly specified. 
In the context of global forecasting models, ensemble methods become especially 
valuable. Global models are trained across many time series simultaneously, 
and while they excel at identifying cross-series patterns, they can also suffer 
from model instability or biases if the chosen architecture fails to generalize 
across heterogeneous series. Combining multiple global models, each with different 
biases, architectures, or feature representations, can help counteract these 
issues by smoothing out individual model errors and reducing the risk of 
overfitting to shared but spurious signals across series \citep{forecombreview}. 
Additionally, ensembles serve as a hedge against model selection uncertainty, 
which is particularly relevant in global settings where the optimal model may 
vary substantially across groups of series. By pooling predictions, ensembles 
offer a form of model diversification that improves generalization, making them 
a powerful and practical tool in modern forecasting systems.

However, ensemble methods are not without drawbacks. One significant limitation 
is the increased computational cost, especially when combining large numbers of 
complex models like deep learning architectures or global forecasting models. 
This cost is magnified when retraining or model tuning is required for each 
component model at multiple forecast origins. Additionally, from a business 
perspective, managing and maintaining multiple models adds complexity to the 
production of forecasts. Moreover, in cases where the component models are very 
similar or highly correlated, the marginal benefit of ensembling diminishes, 
potentially offering little improvement over the best individual model \citep{comb3}.

Therefore, one of the major advantages of adopting a global modeling approach, 
namely scalability, can be offset when ensemble learning is introduced in the
forecasting system. Combining several models, even with 
simple techniques like averaging, inherently increases the computational time, as 
multiple models must be trained, validated, and maintained, producing substantial 
computational overhead. In this sense, ensemble learning may weaken one of the core 
benefits of the global modeling paradigm: reduced training time and resource usage. 
Moreover, since forecasts are typically generated using cloud computing services 
with pay-as-you-go pricing models \citep{occamrazor}, increased computational time 
and resource consumption directly leads to higher forecasting costs for organizations.
These costs may also rise exponentially when frequent retraining strategies are adopted. 
Indeed, it is a widespread practice to update forecasting models whenever new data 
becomes available, often driven by the belief that frequent updates enhance the 
model's ability to adapt to evolving patterns and improve predictive accuracy. 
However, when ensembles are used, all the models within the ensembles must be 
retrained to obtain the final predictions, abruptly increasing the costs of
forecasting. 

This trade-off between performance and computational time in the context of ensemble 
learning raises important questions about the balance between accuracy gains and cost 
efficiency, especially in large-scale forecasting applications. 
However, despite these challenges, ensemble learning remains one of the most widely 
used strategies in the forecasting industry, which further amplifies concerns about 
the long-term sustainability and scalability of forecasting systems. Indeed, from a 
practical perspective, forecast combinations, coupled with frequent retraining, 
has also significant environmental costs, since the energy consumption of model 
training extends beyond direct computational costs, contributing to global energy consumption 
and carbon emissions \citep{greenai2}. Therefore, understanding the cost of ensembling 
forecasting models and the effects of retraining on their performance are of paramount 
importance for advancing more sustainable forecasting practices.


\subsection{Research Question}

We aim to address the question \textit{"Are ensembles always the solution?"}.
Specifically, we focus on the trade-off between accuracy and sustainability, 
in terms of the cost of producing the forecasts, among different types of ensembles 
of global forecasting models. This cost is particularly relevant for ensemble 
models since it becomes exponentially larger with the number of base models used
to produce the combined forecasts.
To this end, we use ten global models as base learners, ranging from traditional 
machine learning methods to advanced deep learning architectures. In addition, we 
simulate eight ensemble approaches, all trained and tested on the two most recent and 
comprehensive retail forecasting datasets: the M5 and VN1 competition datasets.

We also investigate the effects of retraining on the forecasting performance of 
ensembles, compared to that of the base models. In particular, we assess whether 
reducing the frequency of retraining, by avoiding re-estimation of base models as 
new data becomes available, can effectively enhance the cost-accuracy trade-off in 
ensemble forecasting. Hence, we explore a wide range of retraining strategies, 
from continuous retraining to a no-retraining approach, including various intermediate 
periodic retraining schemes to encompass the most practical and effective scenarios.


\subsection{Contributions}

Our contribution is fourfold:

\begin{itemize}

    \item We provide the first comprehensive study of the cost-accuracy trade-off of ensembles
    of global models, using 10 distinct methods, some of the most relevant real-world datasets, 
    and evaluating both point and probabilistic predictions. 

    \item We analyze different ensemble strategies to assess the effects of the ensemble size
    on forecast accuracy and costs.
    
    \item We compare various retraining solutions, such as continuous, periodic, and no 
    retraining, across multiple datasets, to quantify the impact of the retraining frequency 
    on the computational cost of ensemble forecasting.

    \item We provide practical guidelines for organizations and practitioners on assessing 
    when and how frequently to retrain ensemble forecasting models to obtain an effective 
    balance between predictive accuracy and computational cost.    
    
\end{itemize}

By tackling these issues, this paper contributes to both forecasting and machine learning 
communities, offering valuable insights into the trade-offs among accuracy, computational 
efficiency, and sustainability in the use of an ensemble of global forecasting models.


\subsection{Overview}

The rest of this paper is organized as follows. 
After a brief review of related works (Section \ref{sec:literature}), in Section 
\ref{sec:exp_des} we describe the design of the experiment used in our study.
The datasets and their characteristics are presented in \ref{s_sec:data}, while the
global forecasting models and the ensemble strategies are shown in \ref{s_sec:forec_mod}.
The concepts related to rolling origin evaluation and retrain scenario are explained 
in \ref{s_sec:eval_setup}, while the metrics used to assess the accuracy and cost of models 
are discussed in \ref{s_sec:perf_met}. 
In Section \ref{sec:results} we show the empirical findings of our study, 
including forecast accuracy, computing time, and cost analysis
of the different scenarios and across the different ensembles.
Finally, Section \ref{sec:conc} contains our summary and conclusions.


\section{Related works} \label{sec:literature}

The cross-learning (global) approach in time series forecasting has witnessed substantial 
advancements in recent years becoming a central theme in contemporary research. 
Today, global models are commonly used as benchmarks in empirical studies, 
underscoring their growing relevance in the field \citep{investcrosslearn}. 
Additionally, theoretical contributions from \citet{princlocglob} and 
\citet{princlocglob2} have shown that global models can achieve accuracy comparable 
to local models with reduced complexity and without relying on assumptions about data 
similarity. Indeed, global models have proven particularly effective across a variety 
of forecasting domains, especially in retail forecasting \citep{m5acc}, and 
several methods have been proposed to further boost their performance 
\citep{globclus1, globclus2, globaug}.
In particular, ensemble learning is a powerful tool in the forecasters' hands 
offering a reliable strategy to enhance accuracy and robustness by combining the 
strengths of multiple models, not only in the cross-learning context 
\citep{forecombreview}. Relative to point forecasting, many different combination
methods have been proposed: linear combinations with optimal weights \citep{comb1},
performance based weighting \citep{comb2}, criteria-based weighting \citep{comb3}, 
and many different approaches based on stacking, or meta-learning \citep{forecombreview}.
However, time series forecasting competitions have shown that simple ensemble 
strategies, like simple mean or median, are extremely accurate, yet efficient 
\citep{m5comp}. Indeed, simple combination schemes are hard to beat, and the simple 
arithmetic average of predictions with equal weights remains the most widely used 
and surprisingly effective combination rule \citep{forecombpuzzle}. 

Most evaluations of global models emphasize point forecast accuracy, likely due to 
the fact that many machine learning and deep learning algorithms do not natively 
produce probabilistic outputs \citep{m5unc}. However, in applications such as supply 
chain management, quantifying uncertainty is essential, whether through prediction 
intervals, quantiles, or full predictive distributions \citep{retailfor}. 
Forecast combinations in the context of probabilistic predictions are an active area 
of research. Different methods have been proposed, from linear pooling, Bayesian model 
averages, and integral combinations \citep{forecombreview}. However, quantile 
aggregation via simple average has proven to be effective and accurate \citep{ens3, ens4}.

From the perspective of model retraining, the most comprehensive investigation 
in the context of global models has been conducted by \citet{zanotti1}, who found that 
reducing the retraining frequency of global models can lower forecasting costs without 
harming accuracy. Other works, such as \citet{tsupdate1}, explored retraining strategies 
and parameter updates for local exponential smoothing models, while \citet{tsupdate2} 
examined retraining in retail demand forecasting using a limited range of models and 
datasets. Although these studies offer useful insights, direct analysis on the effects
of retraining on ensemble models remains unexplored.

The question of whether ensemble methods are consistently worthwhile is relevant both 
for the forecasting and the broader machine learning community. Indeed, the growing advocacy 
for sustainable AI \citep{greenai2, greenai4} highlights the importance of evaluating 
the environmental and computational costs of model development and maintenance. 
Our study directly addresses this gap by systematically evaluating different ensemble 
strategies' cost-accuracy trade-off. We also assess how different retraining frequencies 
affect the forecasting accuracy across the ensemble models, with the aim of testing 
whether less frequent retraining can be an effective tool to manage this trade-off.
Answering these questions, this research provides both theoretical clarity and practical 
guidelines towards more sustainable forecasting practices.


\section{Experimental design} \label{sec:exp_des} 

This section presents the empirical analysis performed to assess the performance of 
ensemble models and to determine whether reducing the frequency of retraining can 
yield forecasting accuracy comparable to that of a baseline scenario, involving continuous 
retraining, while systematically decreasing the forecasting costs. 
We begin by introducing the datasets used in the experiments, followed by a description 
of the ensemble learning models employed. Finally, we outline the evaluation strategy, 
including the performance metrics, the different retraining scenarios considered, and the 
approach used to assess forecast performance and costs.


\subsection{Datasets} \label{s_sec:data}

For our experiments, we utilized two prominent retail forecasting datasets: 
the M5 and VN1 competition datasets. The M5 competition, part of the 
M-competitions series led by Spyros Makridakis and collaborators, was 
designed to benchmark forecasting methods in a retail demand setting 
\citep{m5comp}. The M5 dataset \citep{m5data} is widely recognized and 
extensively studied, comprising 30,490 daily time series representing unit 
sales of Walmart products. These sales span three main product categories, 
Food, Hobbies, and Household, across ten stores in three U.S. states: 
California, Texas, and Wisconsin. Covering the period from 2011 to 2016, 
the dataset features highly intermittent time series organized hierarchically, 
enabling multi-level forecasting (e.g., individual SKUs, categories, stores, 
and states). It also includes relevant exogenous variables such as prices, 
promotions, and special events like holidays.
The VN1 Forecasting – Accuracy Challenge, launched in October 2024 by Flieber, 
Syrup Tech, and SupChains, marked the first edition of its kind \citep{vn1data}.
This dataset consists of weekly sales data for 15,053 products sold from 2020 to 
2024 by various e-vendors, primarily based in the U.S. Unlike the M5 dataset, 
which contains sales from a single retailer (Walmart) and a limited number of 
physical stores, the VN1 dataset aggregates sales across 328 warehouses operated 
by 46 distinct retailers. To the best of our knowledge, we are among the first 
to benchmark ensemble forecasting models on this dataset. Together, these two datasets 
represent the most recent and comprehensive publicly available collections of 
time series related to retail demand, providing a strong foundation for 
generalizing our findings in the domain of demand forecasting.

In both datasets, we concentrated on the most disaggregated level (SKU $\times$ Store), 
because the potential gains from reduced retraining are greatest at lower 
aggregation levels. To ensure a consistent evaluation strategy 
(see Section \ref{s_sec:eval_setup}), we filtered the series: daily SKUs retained 
at least two years of data (more than 730 observations), while weekly SKUs 
required a minimum of three years (more than 157 observations).


\subsection{Forecasting models} \label{s_sec:forec_mod}

In this study, we focused exclusively on global forecasting methods, as we
used only this category of models as base learners for our ensembles. 
Global approaches have become standard in many industries dealing with 
large-scale time series data, such as retail demand forecasting, where 
predictions must be made for thousands of SKUs \citep{classformeth}. Indeed,
our primary objective is to assess the performance of ensemble models obtained 
by combining the predictions of different global models trained on large 
datasets.

A global forecasting model can be defined as: 

\begin{equation}
    Y_i^h = F(\mathcal{Y}, \Theta) \mbox{,}
    \label{eq:globalmodel}
\end{equation}

where forecasts for a horizon \(h\) for each individual time series \(Y_i\) are 
generated using a single model \(F\) trained on the entire set of time series in 
a dataset \(\mathcal{Y}\). Hence, using the global modeling paradigm, it is 
possible to leverage cross-learning to allow the model to learn shared patterns 
across all time series. Indeed, the model parameters \(\Theta\) are shared 
across all series.

To ensure a comprehensive evaluation of the ensemble models, we included both 
traditional machine learning models and state-of-the-art deep learning techniques
as base learners, chosen for their proven effectiveness in time series forecasting 
and their methodological diversity. 
Additionally, given the importance of evaluating forecasting models against 
established benchmarks in time series analysis, we adopted two well-known 
global models, Linear (Pooled) Regression and Multi-Layer Perceptron, as reference 
methods for comparison throughout our experiments \citep{monashrepo}.

Machine learning models are generally easier to train than deep learning 
approaches but often rely heavily on extensive and careful feature 
engineering to achieve strong forecasting performance \citep{fortrees}. 
In contrast, deep learning models can automatically learn relevant 
features, such as lags or rolling statistics, within their architecture, 
reducing the need for manual preprocessing. However, they are typically 
more challenging to train due to their larger number of hyperparameters, 
which can significantly influence forecast accuracy \citep{esrnn}.
As in \citet{zanotti1}, we implemented simplified feature engineering 
pipelines inspired by the top-performing solutions of the M5 and VN1 
competitions. Our feature set included standard time series features such 
as lags, rolling and expanding means, as well as calendar-related 
variables like year, month, week, and day of the week. We also integrated 
static metadata, such as store, product, category, and location identifiers, 
based on the dataset's characteristics. For the M5 dataset, we further 
included external variables like special events. Hyperparameters were 
chosen based on configurations from leading competition entries when available. 
Otherwise, we used the default values recommended by the respective software libraries.

\subsubsection{Ensemble learning} \label{ss_sec:ens_learn}

Ensemble learning aggregates the predictions of multiple base models to enhance 
forecast accuracy and reliability, especially when different models capture diverse
and complementary patterns in the data. This approach is particularly beneficial in
global forecasting, where the use of a single model can lead to instability or 
overfitting due to the heterogeneous nature of the time series being modeled 
\citep{forecombreview}. To address these challenges, forecast combinations help by 
balancing out the individual biases and variances of the base models.
In a general formulation, an ensemble forecasting model combines the outputs of 
multiple models. Using the same notation introduced before, where \(\mathcal{Y}\) is the set 
of all time series and \(\Theta^{(j)}\) denotes the parameters of the \(j\)-th base 
model \(F^{(j)}\), the ensemble prediction for series \(Y_i\) at horizon \(h\) can 
be defined as:

\begin{equation}
    Y_i^h = G\left( F^{(1)}(\mathcal{Y}, \Theta^{(1)}), F^{(2)}(\mathcal{Y}, \Theta^{(2)}), \dots, F^{(J)}(\mathcal{Y}, \Theta^{(J)}) \right) \mbox{.}
    \label{eq:ensemblemodel}
\end{equation}

Here, \(G\) is the ensemble function (e.g., mean, median, weighted average) that 
combines the forecasts of \(J\) base models.

In our study, we adopted an ensemble approach that combines the models' predictions 
using a simple average. The simple average is a widely used method in forecasting 
to create forecast combinations due to its simplicity, proven effectiveness in 
improving prediction robustness, and is often more accurate than theoretically 
optimal combinations \citep{forecombpuzzle}. 
Therefore, being \(G\) the simple mean, the ensemble forecast for series \(Y_i\) 
at horizon \(h\) becomes:

\begin{equation}
    Y_i^h = \frac{1}{J} \sum_{j=1}^{J} F^{(j)}(\mathcal{Y}, \Theta^{(j)}) \mbox{.}
    \label{eq:ensemblemean}
\end{equation}

Furthermore, since we are not only interested in point forecast accuracy, in the 
context of probabilistic forecasting, ensemble methods can also be applied to 
combine predictive distributions. A common and straightforward approach is to 
average the predicted quantiles across base models \citep{forecombreview}. 
That is, for each desired quantile level (e.g., 0.1, 0.5, 0.9), the ensemble 
forecast simply takes the mean of the corresponding quantiles predicted by the 
individual models. Despite its simplicity, this method is often effective and 
overall seems to be preferred compared to other combination techniques \citep{ens3}.
Moreover, equally weighting quantiles through a simple average yields robust and 
improved forecast results because the error in estimating optimal weights usually 
exacerbates the ensemble predictions \citep{ens4}.
Lastly, simple combination methods have the lowest computational time possible
\footnote{
    Simple combinations, such as the mean or median, require only basic arithmetic 
    operations on the forecast outputs of the base models, avoiding the need for 
    model training or optimization, as it is required by more complex approaches like 
    meta-learning.
}, allowing us to effectively study the associated costs of ensembling without 
any loss of generality.

In our study, we simulate various ensemble strategies to assess, a posteriori, how 
forecast performance over the full test window would have changed if forecasting 
combinations had been used in place of the individual base models. 
Hence, we employed simple mean ensembles to combine the forecasts of multiple 
pools of global models, following two distinct selection criteria. The first strategy, 
which we refer to as \textit{ENSACC}, combines the base models that demonstrated the 
highest individual point forecast accuracy (measured by Equation \ref{eq:rmsse}). 
This approach aims to leverage the strengths of 
top-performing models, under the assumption that their combined output will retain 
strong predictive performance while potentially offsetting individual weaknesses. 
The second strategy, \textit{ENSTIME}, focuses instead on computational efficiency 
by combining the models with the lowest training and inference times. This ensemble 
reflects a pragmatic choice for real-world forecasting systems where computational 
cost is a critical constraint, such as in large-scale retail applications.
For both strategies, we constructed ensembles of increasing size, combining the top 
2, 3, 4, and 5 models according to each criterion. 
This idea of only using a subset of the base learners in ensembling is essentially a variant 
of model pooling, as defined by \citet{ens5}. 
This tiered design allows us to assess how forecast accuracy and computational 
efficiency evolve as additional models are added to the ensemble. Limiting the maximum 
ensemble size to five models (out of eight or ten, depending on the dataset) reflects 
a balance between potential accuracy gains and the diminishing returns or increased 
complexity often observed with larger ensembles \citep{forecombreview}. This systematic 
approach enables a thorough evaluation of the trade-off between accuracy and 
computational cost in ensemble-based global forecasting.

All models were implemented in Python using Nixtla's framework \citep{nixtla}. 
Specifically, the \textit{mlforecast} library was used to train the machine learning 
models, while \textit{neuralforecast} was employed for efficiently training the deep 
learning models.


\subsection{Evaluation strategy} \label{s_sec:eval_setup}

In this section, we introduce the concepts of retraining scenarios and
rolling window forecasting that we used in our experiment.

\subsubsection{Retrain scenario} \label{ss_sec:ret_scn}

Following \citet{zanotti1}, we evaluated multiple retraining scenarios, 
or retrain windows. A retrain scenario \(r\) denotes a positive integer 
indicating how often the model is updated, or retrained. Specifically, 
\(r\) defines the number of new observations that must be collected before 
retraining occurs. These scenarios are tailored to the frequency of each 
dataset, which in turn determines the forecast horizon and business review 
cycles. Table \ref{tab:retrain_scenario} summarizes the selected 
retraining sets, training and test sizes, and forecast horizons.

\begin{table}[ht]
    \caption{Retraining set, train and test size, frequency, forecast horizon,
    and step size for each dataset. The setup is based on the respective time 
    series frequency.}
    \centering
    \begin{tabular}{lll} 
    \toprule
    & \textbf{M5} & \textbf{VN1} \\
    \midrule
    Frequency (\(f\)) & daily (7) & weekly (52) \\
    Train size (\(n\)) & $\geq 730$ & $\geq 157$ \\
    Test size (\(T\)) & 364 & 52 \\
    Horizon (\(h\)) & 28  & 13 \\
    Step size (\(p\)) & 1 & 1 \\
    Retrain set (\(r\)) & \{7, 14, 21, 30, 60, 90, 120, 150, 180, 364\} & \{1, 2, 3, 4, 6, 8, 10, 13, 26, 52\} \\
    \bottomrule
    \end{tabular}
    \label{tab:retrain_scenario}
\end{table}

For example, \(r=7\) for daily data implies weekly retraining. Each list contains 
ten scenarios to ensure a broad yet computationally feasible exploration.
The scenario with \(r=1\), known as continuous retraining, is the most computationally 
intensive and typically the most accurate, as the model always uses the most recent 
data. Hence, we adopt it as our benchmark for both accuracy and cost. However, for 
daily data, we consider \(r=7\) (weekly retraining) as the practical benchmark, since 
daily updates are uncommon in real-world settings.
Conversely, the no-retraining scenario \(r=T\), where \(T\) is the length of the test 
set, fits the model only once, and uses it for the entire forecasting horizon, 
minimizing computational load but likely yielding the lowest accuracy. All intermediate 
values \(1<r<T\) represent periodic retraining strategies, where both forecast accuracy 
and computational cost are expected to decrease as \(r\) increases.

\subsubsection{Rolling window evaluation} \label{ss_sec:roll_win}

Out-of-sample evaluation is essential in time series forecasting to assess a 
model’s ability to generalize to unseen data, especially given the potential 
for structural changes or unanticipated shifts in future values \citep{tseval}.
Among evaluation strategies, rolling origin evaluation is widely recognized as 
the most appropriate approach \citep{tscrossval}. This method systematically 
assesses forecast accuracy over multiple iterations by simulating repeated 
forecasting cycles. The procedure starts by splitting the series into a 
training and a test set, maintaining the chronological order. At each step, 
the model is trained on the current training set and used to predict the next 
\(h\) observations. The forecast origin is then shifted forward by a fixed step 
(the step size \(p\)), and the process is repeated, either with a growing 
(expanding) or fixed-size training window. Performance metrics (see Section 
\ref{s_sec:perf_met}) are averaged over all iterations to provide a robust estimate 
of forecasting accuracy.

Compared to fixed origin evaluation, which offers only a single evaluation point, 
rolling origin evaluation provides a more robust assessment by capturing performance 
across varying conditions such as seasonal shifts, level changes, or trend 
evolutions \citep{tscrossval}. This is particularly beneficial in dynamic settings 
like retail or supply chain management, where models need to adapt to frequent changes.
The setup is flexible: the training window can either expand to include all available 
past data or remain fixed to a specified length. Most practitioners favor the 
expanding window, especially when time series are short, as it leverages the full 
data history \citep{fortheopract}. In our study, we adopt the expanding window 
strategy to reflect realistic business forecasting practices better and to accommodate 
short series, such as those in the weekly VN1 dataset.
Table \ref{tab:retrain_scenario} outlines the key parameters used in our experiments.
In particular, we used a test set of one complete year to avoid results being 
affected by intra-year fluctuations, and we set the forecast horizon based on 
the type of business decisions that usually daily and weekly forecasts support.
Lastly, we adopted a step size of one to maximize the number of evaluations 
across each retraining scenario.


\subsection{Evaluation metrics} \label{s_sec:perf_met}

Evaluating the accuracy of point forecasts in time series analysis remains 
a debated topic. Although numerous metrics exist to assess model performance, 
there is no clear consensus in the literature on which metric is superior 
\citep{forevalds}. Given that our focus is on SKU-level demand forecasting, 
where data is typically intermittent, error metrics based on absolute or 
percentage deviations are suboptimal, as they emphasize the median rather 
than the full distribution \citep{evalbestacc}. Furthermore, due to scale 
variations across series, employing a scale-independent metric is essential.
To address these concerns, we adopted the Root 
Mean Squared Scaled Error (RMSSE), introduced by \citet{evalmeasacc}, as 
our primary point forecast accuracy measure. RMSSE is calculated as:

\begin{equation}
   \text{RMSSE} = \sqrt{\frac{\frac{1}{h} \sum_{t=n+1}^{n+h} 
   (y_t - \hat{y}_t)^2}{\frac{1}{n-s} \sum_{t=s+1}^{n} 
   (y_t - y_{t-s})^2}} \mbox{.}
   \label{eq:rmsse}
\end{equation}

This metric compares the mean squared error of a forecast to that of a 
seasonal naïve benchmark, providing a relative performance measure. It was 
the official evaluation metric used in the M5 competition (with \(s = 1\), 
hence using the naive model as a benchmark) \citep{m5comp}. Lower RMSSE 
values indicate more accurate forecasts.

Beyond point accuracy, we also assessed the probabilistic quality of 
forecasts across different retraining scenarios. To evaluate the quality 
of the estimated quantiles, we used a scaled version of the Quantile Loss 
(SQL), also known as Pinball Loss, and its average form, the scaled 
Multi-Quantile Loss (SMQL):

\begin{equation}
    \text{SQL} = \frac{1}{h} \frac{\sum_{t = n+1}^{n+h} 
    \left( q \cdot (y_t - \hat{y}_t) \cdot \mathbb{I}_{y_t \geq \hat{y}_t} 
    + (1 - q) \cdot (\hat{y}_t - y_t) \cdot \mathbb{I}_{y_t < \hat{y}_t} 
    \right)}{\frac{1}{n-s} \sum_{t=s+1}^{n} |y_t - y_{t-s}|} \mbox{,}
    \label{eq:sql}
\end{equation}

\begin{equation}
    \text{SMQL} = 
    \frac{1}{|\mathcal{Q}|} \sum_{q \in \mathcal{Q}} \text{SQL}(q) \mbox{,}
    \label{eq:smql}
\end{equation}

As a proper scoring rule, QL enables a rigorous evaluation of probabilistic 
forecasts \citep{evalcountdata}. The scaling factor is the in-sample,
one-step-ahead mean absolute error of the seasonal Naive model.
The scaled MQL, particularly in its weighted form and with \(s = 1\), 
was the principal metric of the M5 Uncertainty competition \citep{m5unc}.
For daily data we used \(s = 7\), because \(s = 1\) is arguably an irrelevant 
benchmark in daily retail data, which typically exhibits strong day-of-week 
seasonality \citep{retailfor}. Instead, we set \(s = 1\) for weekly data.

For our analysis, we considered a total of 14 different quantiles, \(q\), 
of probability levels \{0.005, 0.025, 0.050, 0.100, 0.150, 0.200, 0.025, 0.750, 
0.800, 0.850, 0.900, 0.950, 0.975, 0.995\}. Therefore, we studied the 50\% 
60\%, 70\%, 80\%, 90\%, 95\%, and 99\% symmetric prediction intervals. 
The lower intervals (e.g., 50\%, and 60\%) describe the forecast 
center, while the upper intervals (90\% and above) are crucial for assessing tail 
risks, key to managing safety stock in retail settings \citep{evalinventory}. 
These quantiles offer a well-rounded view of forecast uncertainty.

In addition to accuracy, we explicitly evaluated the computational cost of generating 
forecasts by measuring Computing Time (CT), defined as the number of seconds needed to 
train a model and produce \(h\) step-ahead forecasts \citep{zanotti1}. CT directly 
reflects forecasting costs in real-world applications, especially when using cloud-based 
infrastructure with pay-as-you-go pricing \citep{tsupdate1}. Thus, CT was recorded for 
each model across all retraining scenarios, with lower CT values indicating more 
efficient forecasting.

To facilitate consistent comparisons, results for all evaluation metrics were normalized 
relative to the baseline (continuous) retraining scenario for each dataset frequency. 
To statistically assess performance differences across scenarios, we applied the 
Friedman-Nemenyi test for multiple comparisons \citep{testfriednem}.

All experiments were conducted using a Microsoft Azure NC6s\_v3 cloud instance running 
Ubuntu 24, equipped with 1 GPU, 6 CPU cores, and 112 GB of memory.


\section{Results and discussion} \label{sec:results}

This section presents and interprets the empirical findings of our study, highlighting 
the interplay between accuracy, probabilistic performance, computing time, and total 
cost across different retraining strategies and ensemble configurations. We draw 
insights from both the M5 and VN1 datasets to identify consistent patterns and 
dataset-specific nuances. 

\begin{table}
    \centering
    \caption{Updated forecasting performance and computational cost for each method 
    across datasets. RMSSE, SMQL, and CT (in seconds). Minimum RMSSE and SMQL values 
    per column are highlighted in bold. The baseline scenarios are \(r = 7\) for the 
    M5 dataset and \(r = 1\) for the VN1 dataset.
    }
    \label{tab:combined_metrics}
    \begin{tabular}{llrrrrrr}
    \toprule
    \multirow{2}{*}{\textbf{Type}} & \multirow{2}{*}{\textbf{Method}} & \multicolumn{3}{c}{\textbf{M5}} & \multicolumn{3}{c}{\textbf{VN1}} \\
    \cmidrule(lr){3-5} \cmidrule(lr){6-8}
     &  & \textbf{RMSSE} & \textbf{SMQL} & \textbf{CT} & \textbf{RMSSE} & \textbf{SMQL} & \textbf{CT} \\
    \midrule
    \multirow{5}{*}{ML}
     & LR         & 0.760 & 0.269 & 11373  & 6.535 & 4.005 & \textbf{236} \\
     & RF         & --    & --    & --     & 1.871 & 1.220 & 24862 \\
     & XGBoost    & \textbf{0.739} & 0.246 & 15417  & 1.898 & 1.503 & 530 \\
     & LGBM       & 0.755 & 0.247 & 44429  & 3.540 & 2.318 & 7824 \\
     & CatBoost   & 0.926 & 0.280 & \textbf{10424}  & 5.355 & 3.889 & 805 \\
    \midrule
    \multirow{5}{*}{DL}
     & MLP        & 0.804 & 0.264 & 17584  & 1.554 & \textbf{0.806} & 962 \\
     & LSTM       & --    & --    & --     & 1.960 & 0.892 & 1284 \\
     & TCN        & 0.847 & 0.271 & 33364  & 1.960 & 0.891 & 1127 \\
     & NBEATSx    & 0.799 & 0.263 & 21226  & 1.704 & 0.991 & 1244 \\
     & NHITS      & 0.812 & 0.267 & 21969  & 1.705 & 0.992 & 1251 \\
    \midrule
    \multirow{4}{*}{ENSACC}
     & Ens2A      & 0.741 & 0.244 & 59846  & \textbf{1.478} & 0.826 & 2205 \\
     & Ens3A      & 0.741 & 0.248 & 71219  & 1.523 & 0.867 & 3456 \\
     & Ens4A      & 0.743 & \textbf{0.241} & 92445  & 1.530 & 0.926 & 28318 \\
     & Ens5A      & 0.748 & 0.242 & 110029 & 1.550 & 1.013 & 28849 \\
    \midrule
    \multirow{4}{*}{ENSTIME}
     & Ens2T      & 0.798 & 0.257 & 21797  & 5.694 & 4.388 & 766 \\
     & Ens3T      & 0.762 & 0.249 & 37214  & 5.286 & 3.972 & 1572 \\
     & Ens4T      & 0.751 & 0.242 & 54797  & 4.447 & 3.420 & 2533 \\
     & Ens5T      & 0.750 & 0.243 & 76024  & 3.664 & 2.832 & 3660 \\
    \bottomrule
    \end{tabular}
\end{table}

Table \ref{tab:combined_metrics} summarizes the performance and computational time of 
all forecasting methods evaluated in this study, across both the M5 and VN1 datasets. 
The models are grouped into four categories: Machine Learning (ML), Deep Learning (DL), 
ENSACC (ensembles optimized for accuracy), and ENSTIME (ensembles optimized for 
computational efficiency). For each method, the table reports three key metrics per dataset:
RMSSE (Root Mean Squared Scaled Error) to assess point forecast accuracy, SMQL 
(Scaled Multi-Quantile Loss) to evaluate probabilistic forecast performance, and CT 
(Computing Time, in seconds) to reflect the computational time under a cloud-based setting.
Overall
\footnote{
    The overall results are derived from the baseline scenario, that is \(r = 7\) 
    for M5 and \(r = 1\) for VN1, as this setting is regarded as standard in 
    both theoretical and practical applications.
}, the models evaluated achieved better absolute performance on the M5 dataset 
than on VN1. Several factors may explain this difference, including the larger dataset 
size, the higher frequency of the time series in M5, the availability of rich external 
regressors (e.g., promotions, special events), and the presence of well-established 
benchmark hyperparameter settings, many of which were not available for VN1 during model 
training.

Analyzing the point forecasting accuracy, the RMSSE confirms the 
effectiveness of ensemble methods in both datasets. In general, 
ENSACC ensembles (those optimized for accuracy) consistently outperform individual base 
models. The accuracy improvements, however, show diminishing returns beyond three or four 
models, aligning with the well-documented phenomenon in ensemble literature where the 
marginal gain from adding more models decreases. Notably, while ENSACC ensembles dominate 
in accuracy, ENSTIME ensembles (optimized for computational efficiency) still provide 
competitive results in the M5 dataset, demonstrating that even time-efficient combinations 
can retain good predictive power. Interestingly, ENSTIME ensembles show increasing returns 
in accuracy as more models are added: their performance consistently improves with each 
additional component. Indeed, in contrast to ENSACC ensembles, where combining just two 
models is often sufficient to achieve optimal performance, ENSTIME combinations benefit 
from larger ensemble sizes, with forecast accuracy continuing to improve as more low-cost 
models are included (see Supplementary material for significance tests).

A similar pattern emerges for probabilistic accuracy evaluated using Scaled 
Multi-Quantile Loss (SMQL). Compared to base models, ensembles 
achieve superior uncertainty quantification, benefiting from the diversity in quantile 
predictions among different models. Simple averaging across quantiles proves to be a highly 
effective strategy. Also, in this case, ENSACC forecast combinations exhibit diminishing 
returns in accuracy as more models are added. On the contrary, increasing the number of 
models in the ensemble based on computational efficiency leads to consistently improved 
probabilistic forecasting performance, which may even outperform the ENSACC combinations. 
ENSTIME ensembles achieve similar probabilistic forecasting performance on 
the M5 dataset compared to all accuracy-based combinations, while their performance is
worse on the VN1 dataset (see also Supplementary material). 
In contrast, for ENSACC ensembles, combining more than two base models does not 
yield additional improvements, confirming once again that a small, high-performing subset is 
sufficient in that case.

Unsurprisingly, ensemble models require significantly more computing time than individual 
models, with algorithms like LGBM and RF contributing disproportionately to this overhead. 
ENSACC ensembles are especially time-intensive, as they often include the most 
computationally demanding models. In contrast, ENSTIME ensembles maintain a more 
manageable runtime profile by design, though they still incur a higher computational cost 
than their individual components due to the cumulative effect of combining multiple models.

\begin{figure}
    \centering
    \includegraphics[scale=0.55]{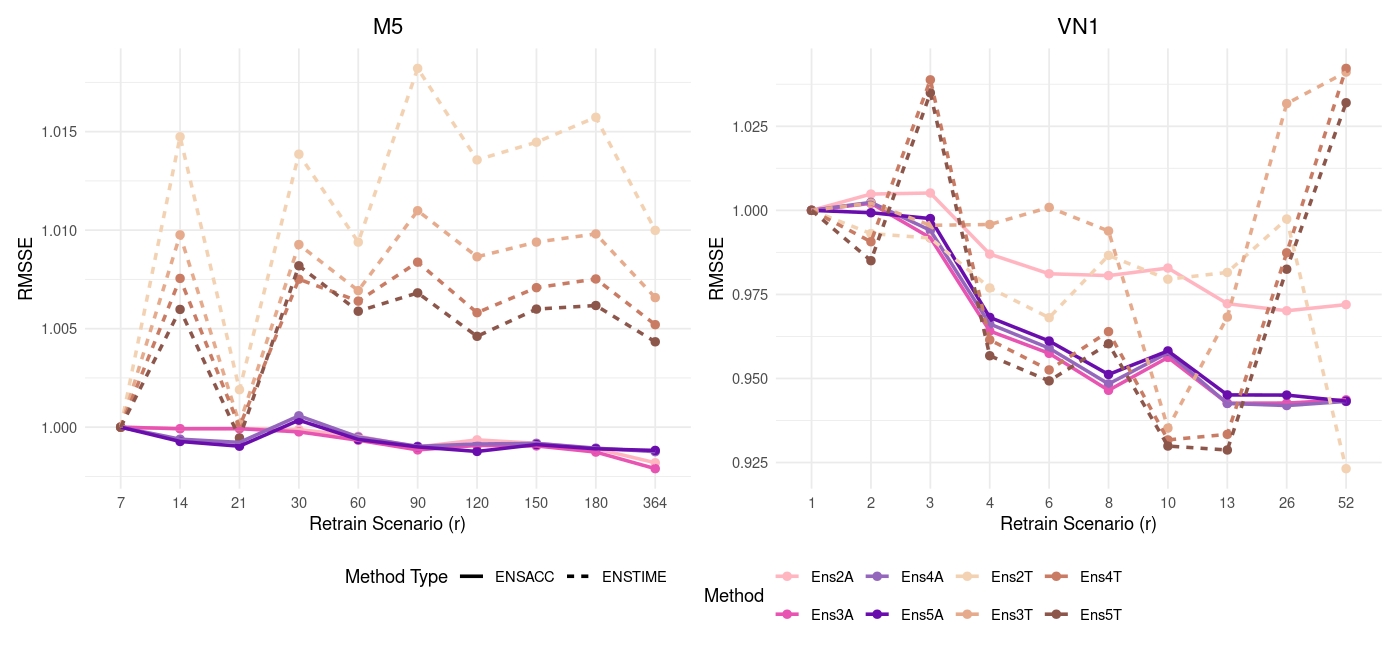}
    \caption{RMSSE results for each ensemble and 
    retrain scenario combination in relative terms 
    with respect to the baseline scenario, \(r = 7\) 
    for the M5 dataset and \(r = 1\) for the VN1 dataset.}
    \label{fig:rmsse_plot}
\end{figure}

Figure \ref{fig:rmsse_plot} presents the point forecast accuracy of each ensemble model 
across various retraining scenarios for both the M5 and VN1 datasets. To facilitate 
comparison, the results are expressed in relative terms with respect to the baseline 
retraining scenarios (\(r = 7\) for M5 and \(r = 1\) for VN1).
The RMSSE trajectories of the ENSACC ensembles exhibit remarkable stability across 
retraining periods. In the M5 dataset, their accuracy remains virtually unchanged 
regardless of retraining frequency, while in the VN1 dataset, performance even improves 
in some scenarios. For most ensemble models, particularly under low-frequency 
retraining (i.e., periodic updates), performance is nearly indistinguishable from the 
baseline. Although slight accuracy degradation is observed for the ENSTIME ensembles 
under less frequent retraining, it remains modest, below 2\% in M5 and under 3\% in VN1, 
even in the no-retraining condition. 
These findings build upon and extend the results of \citet{zanotti1}, suggesting that 
reducing retraining frequency does not significantly harm the point forecast accuracy 
of global ensemble models. This robustness likely stems from the relative stability of 
the datasets, which do not exhibit substantial structural changes or concept drift. 
Under such conditions, base models continue to capture underlying demand patterns 
effectively over time, and ensemble combinations further enhance stability as retraining 
becomes less frequent.

\begin{figure}
    \centering
    \includegraphics[scale=0.55]{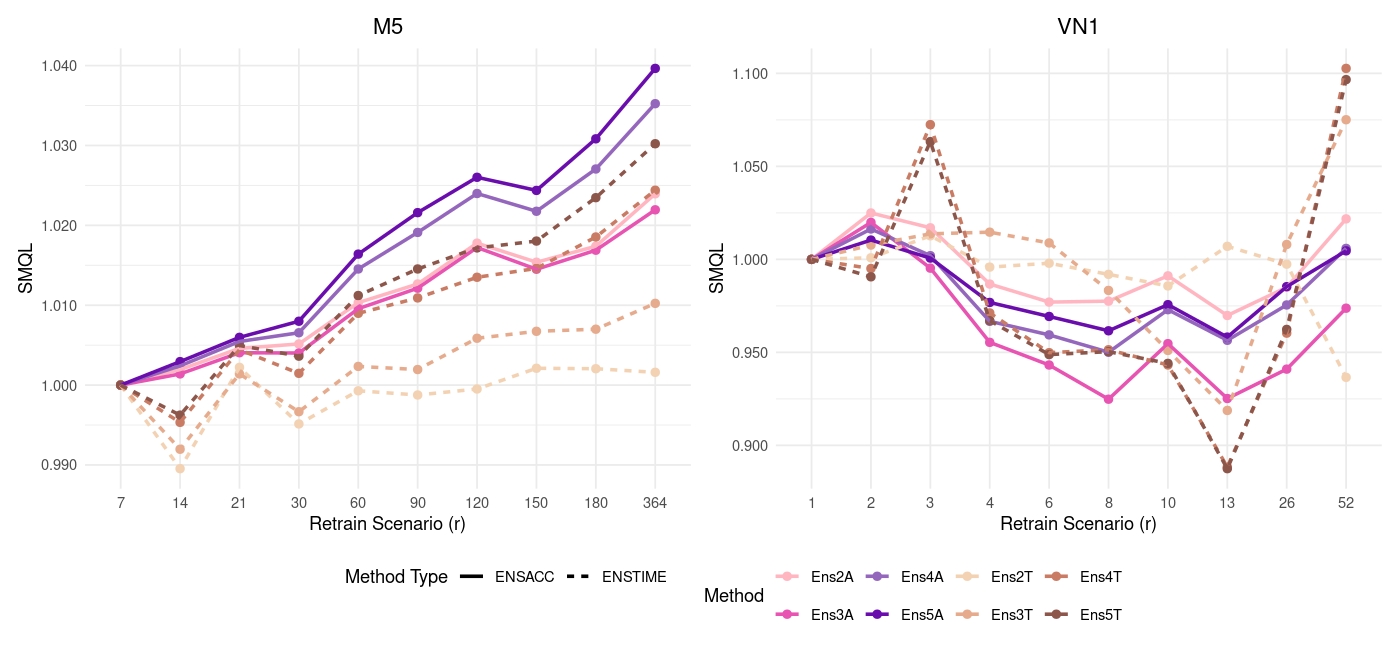}
    \caption{SMQL results for each ensemble and 
    retrain scenario combination in relative terms 
    with respect to the baseline scenario, \(r = 7\) 
    for the M5 dataset and \(r = 1\) for the VN1 dataset.}
    \label{fig:smql_plot}
\end{figure}

Similarly, Figure \ref{fig:smql_plot} illustrates the relative accuracy of the ensemble 
models in a probabilistic forecasting context, as measured by the Scaled Multi-Quantile Loss 
(SMQL). For the M5 dataset, we observe a clear trend where probabilistic accuracy 
decreases as retraining becomes less frequent, indicating that regular updates are 
beneficial for maintaining high-quality uncertainty estimates. This pattern holds across 
the different ensemble strategies. For ENSACC ensembles the decline in accuracy 
is minimal for frequent retraining scenarios and becomes slightly more noticeable at higher 
levels of retraining, though it remains within 4 percentage points. 
ENSTIME combinations, instead, show even higher performance, in particular in the case of 
the smallest ensembles. 
In contrast, the VN1 dataset reveals a different behavior. Here, the relationship between
retraining frequency and probabilistic accuracy follows a near-convex pattern. Performance
initially improves with less frequent retraining, peaking around \(r = 8\), and begins to 
decline thereafter. This suggests that very frequent updates may not be necessary for 
probabilistic performance in stable, lower-frequency datasets like VN1. 
Interestingly, the rate of performance deterioration with increased retraining window is
comparable across both ensemble strategies in VN1, indicating that the marginal impact of
retraining frequency is relatively uniform across the proposed ensemble techniques.
Also these results on the probabilistic performance of ensembles are comparable to those 
obtained by \citet{zanotti1} on the base models.

The Friedman-Nemenyi tests, used to compare the accuracy produced by the same model over 
different retraining scenarios, confirm the above results (see Supplementary material).
Specifically, in the evaluation of both point forecast accuracy and probabilistic 
forecast accuracy, some level of periodic retraining is at least as good as the 
continuous retraining scenario.

\begin{figure}[ht]
    \centering
    \includegraphics[scale=0.55]{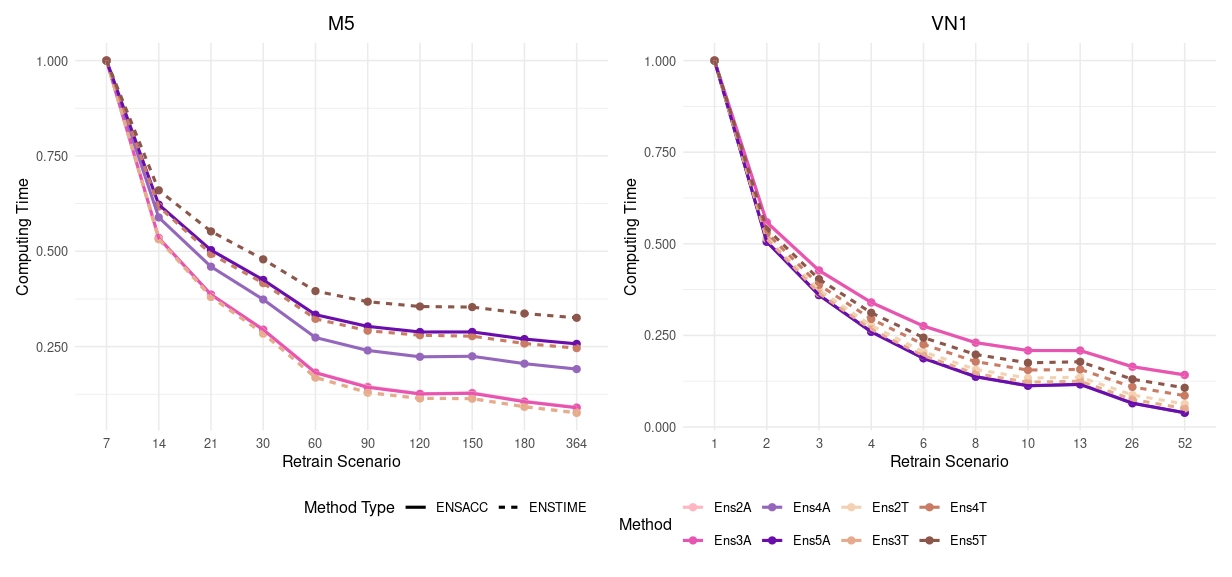}
    \caption{CT results for each ensemble and 
    retrain scenario combination in relative terms 
    with respect to the baseline scenario, \(r = 7\) 
    for the M5 dataset and \(r = 1\) for the VN1 dataset.}
    \label{fig:ct_plot}
\end{figure}

Figure \ref{fig:ct_plot} illustrates the relative computational time (CT) across 
different retraining scenarios for both the M5 and VN1 datasets. As anticipated, 
CT decreases sharply as the retraining interval increases, with the reduction 
following an approximately exponential pattern. This effect is more pronounced 
in the VN1 dataset, likely due to its smaller size compared to M5. Specifically, 
moving from the baseline retraining scenario to the first periodic setting 
(\(r = 14\)) results in a roughly 33\% reduction in computing time for M5, while 
for VN1, CT is nearly halved.

As dataset size increases, the marginal gains in CT from reducing retraining 
frequency appear to depend more on the number of models in the ensemble (ensemble 
depth) than on the ensemble type (ENSACC vs. ENSTIME). Simpler ensembles composed 
of only 2 or 3 models benefit most from reduced retraining, exhibiting steeper 
declines in CT. Moreover, ENSACC and ENSTIME ensembles show very similar CT profiles 
when matched by depth, suggesting that computational efficiency in this context is 
more a function of ensemble size than selection strategy. This distinction is less 
pronounced in the VN1 dataset, given its smaller scale and shorter time series. 
These findings have direct implications for forecasting cost management, emphasizing 
the importance of considering retraining frequency and ensemble size when selecting 
models for large-scale operational use.

Overall, the evidence from Figures \ref{fig:rmsse_plot}, \ref{fig:smql_plot}, and 
\ref{fig:ct_plot}, along with the statistical significance tests, indicates that 
reducing the retraining frequency of ensemble models does not negatively impact 
their predictive accuracy. At the same time, it lowers the computational time 
required to generate forecasts. However, effectively leveraging this time reduction 
depends on carefully considering the ensemble size. As more models are included in 
the ensemble, the relative benefit of periodic retraining diminishes compared to 
using only a single model as in \citet{zanotti1}. Consequently, both the retraining 
frequency and the ensemble depth play a critical role in managing computational 
efficiency, and ultimately the cost, of ensemble-based forecasting systems.

\begin{figure}
    \centering
    \includegraphics[scale=0.43]{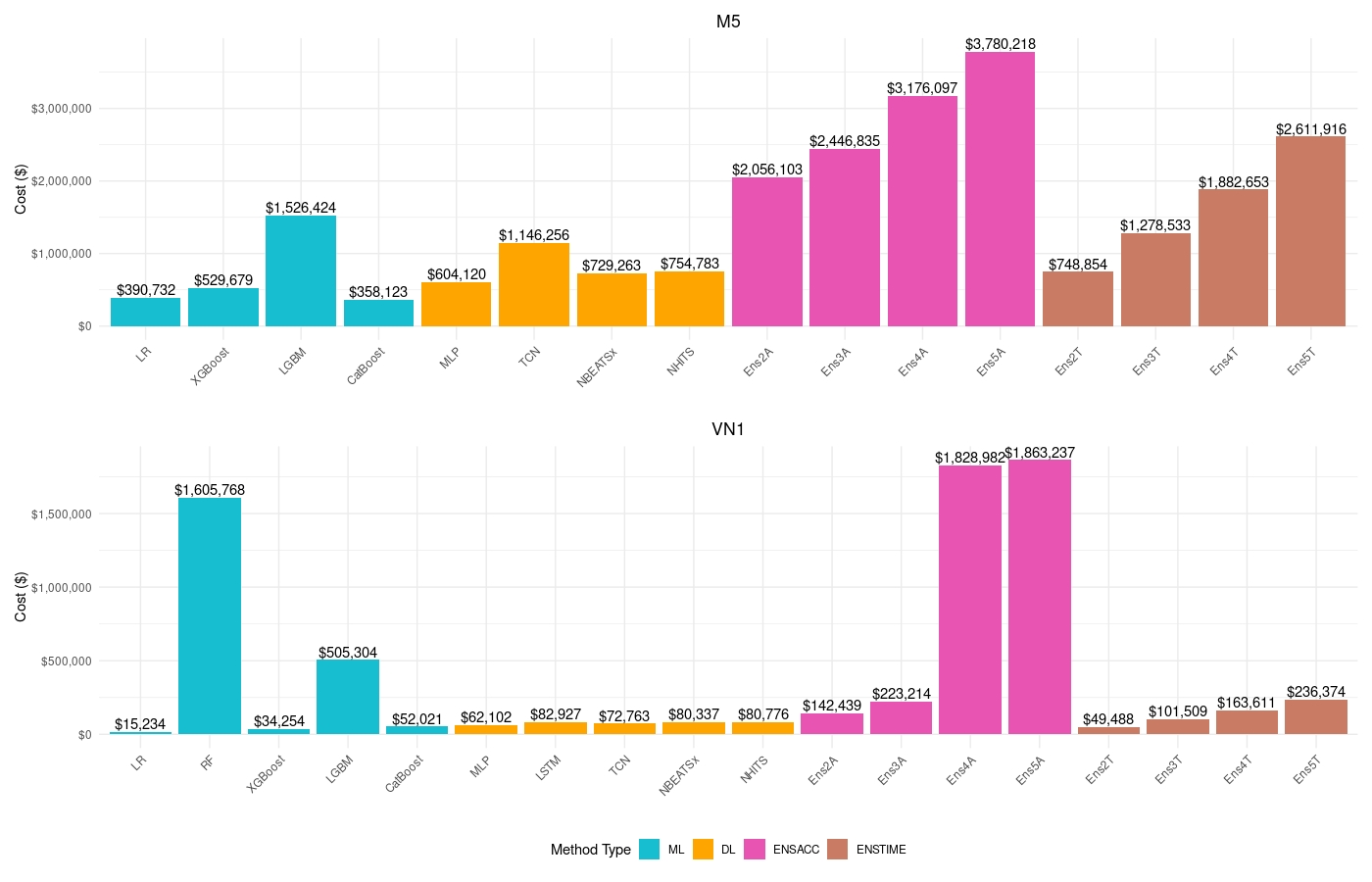}
    \caption{Cost analysis. Overall cost results for the M5 and VN1 datasets.}
    \label{fig:overall_cost_plot}
\end{figure}

\begin{figure}
    \centering
    \includegraphics[scale=0.5]{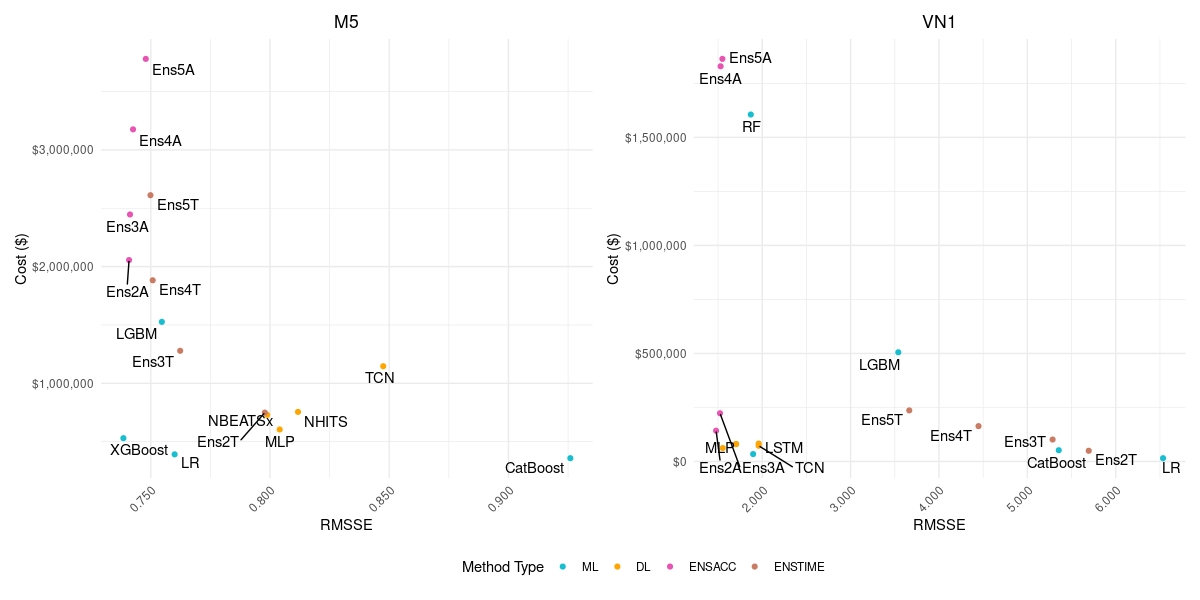}
    \caption{Cost analysis. RMSSE vs Cost (\$) for the M5 and VN1 datasets.}
    \label{fig:rmsse_vs_cost_plot}
\end{figure}

\begin{figure}
    \centering
    \includegraphics[scale=0.5]{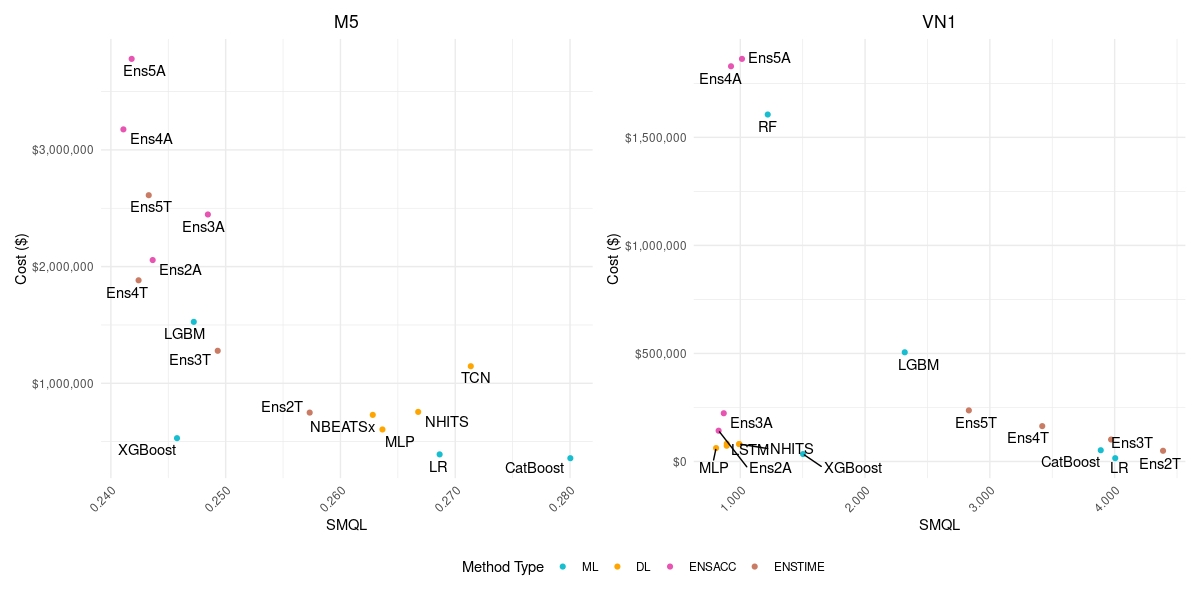}
    \caption{Cost analysis. SMQL vs Cost (\$) for the M5 and VN1 datasets.}
    \label{fig:smql_vs_cost_plot}
\end{figure}

\begin{figure}
    \centering
    \includegraphics[scale=0.4]{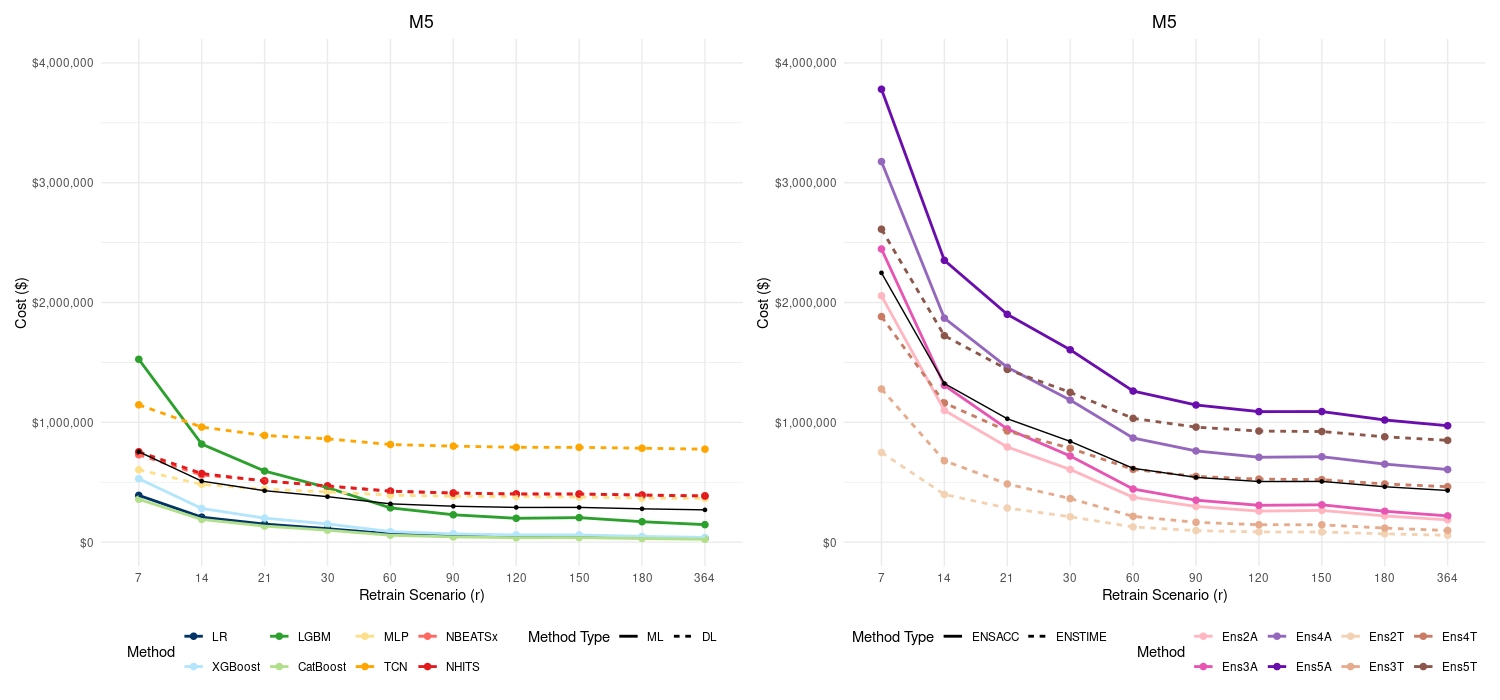}
    \caption{Cost analysis. Comparison between base models and ensembles
    for the M5 dataset. 
    Each method and retrain scenario combination are shown 
    in relative terms with respect to the baseline scenario, 
    \(r = 7\).}
    \label{fig:m5_cost_plot}
\end{figure}

As previously noted, computational time can be directly translated into monetary 
cost for organizations. In line with \citet{forsubopt, occamrazor, zanotti1}, we 
adopted standard pricing assumptions for cloud computing services to estimate both 
the overall cost of forecasting and the specific costs associated with each retraining 
scenario for the base global models and each ensemble
\footnote{
    The costs have been normalized by the number of SKUs in each dataset to allow comparisons.
}.
Figures \ref{fig:overall_cost_plot}, \ref{fig:rmsse_vs_cost_plot}, \ref{fig:smql_vs_cost_plot},
and \ref{fig:m5_cost_plot} illustrate the estimated forecasting 
costs for a large-scale retailer, assuming a computing service rate of \$3.5 per hour, with 
200,000 unique SKUs and 5,000 stores.
The total forecast production cost (Figure \ref{fig:overall_cost_plot}), 
computed by combining training time and forecast generation 
under a cloud-computing pricing model, highlights the trade-off central to our study.
ENSACC ensembles incur the highest costs, rapidly reaching millions of dollars under 
frequent retraining. In contrast, ENSTIME ensembles offer a more sustainable alternative, 
reducing costs by over 50\% in many cases. 
Moreover, as dataset size increases, the trade-off between forecast accuracy and cost becomes 
significantly more pronounced. This is clearly illustrated in Figures \ref{fig:rmsse_vs_cost_plot} 
and \ref{fig:smql_vs_cost_plot}. For the VN1 dataset, ENSACC methods offer noticeable 
improvements in both point and probabilistic forecasting accuracy compared to most individual 
base models, while incurring only a modest increase in cost. In this context, there is little 
reason to favor ENSTIME ensembles over ENSACC ones, as the cost difference is minor and the 
performance gain from accuracy-based combinations is evident.
However, the situation is quite different for the M5 dataset. Here, ensemble models, particularly 
those optimized for accuracy, are significantly more expensive than individual models. Additionally, 
the performance advantage of ENSACC ensembles over the most accurate base models is minimal in 
terms of both RMSSE and SMQL. In contrast, ENSTIME ensembles can achieve comparable levels of 
accuracy with substantially lower additional cost, making them a more attractive option. It is also 
clear that increasing the number of models in an ensemble does not necessarily lead to proportional 
gains in forecasting accuracy, which raises questions about the cost-effectiveness of larger 
combinations.
Notably, forecast combinations that disregard the computational efficiency of their base models can 
lead to excessive and unjustified costs. This is particularly evident in the case of Ens4A and Ens5A. 
For instance, on the VN1 dataset, where the inclusion of Random Forest (a highly expensive 
model in this context) results in disproportionately high forecasting costs with only marginal 
accuracy improvements.

Finally, \ref{fig:m5_cost_plot} shows the comparison between base 
models and ensembles across all retraining scenarios. As expected, forecasting costs decrease
exponentially with less frequent retraining. Across both datasets, ENSTIME ensembles 
consistently incur lower costs than their ENSACC counterparts. Moreover, the cost-saving 
benefits of reduced retraining frequency become more pronounced as dataset size increases. In 
the VN1 dataset, cost differences between retraining scenarios remain relatively small, even 
for ensemble models, suggesting limited sensitivity to retraining frequency (see Supplementary 
material). In contrast, the M5 dataset exhibits steady, incremental reductions in cost as 
retraining becomes less frequent.
For example, in the M5 setting, the average cost of continuous retraining for ensemble models 
is approximately \$2,250,000, more than triple the average cost of individual base models. When 
retraining is entirely eliminated, this figure drops to around \$500,000, representing a cost 
reduction of over 75\%. However, even in the no-retraining scenario, ensemble models remain 
roughly twice as expensive as the average base model.
Interestingly, the cost gap between ENSACC and ENSTIME ensembles narrows for smaller ensembles 
(e.g., those with two or three base models) as retraining frequency decreases. This indicates 
that with less frequent retraining, accuracy-based combinations (ENSACC) become increasingly 
comparable to efficiency-based ones (ENSTIME) in terms of forecasting cost. Thus, infrequent 
retraining can help align performance-optimized ensemble strategies with more budget-conscious 
forecasting objectives.

One might argue that, for a large retailer, these costs (and the potential savings) 
are relatively small. However, it's important to emphasize that the cost reductions 
achieved through a combination of less frequent retraining, smaller ensemble sizes, 
and more efficient model combinations typically come with no loss in forecasting 
accuracy, particularly for point forecasts. This makes such strategies both economically 
and operationally attractive.
These results further support our findings. ENSACC ensembles dominate in accuracy but 
at a significant computational premium. ENSTIME ensembles represent a middle ground. 
Importantly, ensembles of just two or three models (e.g., ENS2A, ENS3T) often 
achieve near-optimal performance, providing a compelling trade-off between complexity, 
cost, and accuracy, and lowering the retraining frequency can be a good strategy to improve 
this trade-off even in the context of ensemble learning.
This comprehensive analysis reinforces the core message of the article: ensemble methods 
offer strong accuracy benefits, but their cost can be substantial, especially under 
frequent retraining. Balancing accuracy and sustainability requires deliberate model 
selection and retraining design, with lightweight ensembles, like ENSTIME, being a 
possible solution in large-scale forecasting systems.


\section{Conclusions} \label{sec:conc}

This study investigated the cost-effectiveness of ensemble learning in global time 
series forecasting, focusing on the interplay between forecast accuracy, computational 
efficiency, and retraining strategies. We evaluated ten global forecasting models, 
spanning both classical machine learning and deep learning techniques, alongside eight 
ensemble configurations, across two large-scale and industry-relevant datasets: the 
M5 and VN1 retail forecasting datasets. By systematically comparing different ensemble 
sizes, selection criteria (accuracy-based vs. efficiency-based), and retraining 
frequencies, our aim was to assess whether the accuracy gains of ensemble methods 
justify their additional computational cost, particularly in large-scale operational 
settings.

Our findings showed that ensemble models consistently improve both point and probabilistic 
forecasting performance over most individual base models. This improvement is also  
relevant for probabilistic accuracy, where ensembles provide better uncertainty 
quantification through the aggregation of diverse quantile predictions. The results 
support the notion that ensemble learning offers a robust mechanism for improving 
predictive reliability and mitigating individual model weaknesses, even in the context 
of global modeling. However, these benefits come at a cost. Ensemble methods are 
substantially more computationally intensive than single-model approaches, particularly 
when composed of complex or resource-heavy algorithms such as LightGBM or Random Forest. 
These computational demands translate directly into higher forecasting costs under 
cloud-based computing environments.

Our analysis further distinguished between two ensemble design philosophies: ENSACC 
(accuracy-based) and ENSTIME (efficiency-based). While ENSACC ensembles generally deliver 
higher accuracy, especially when composed of two or three strong base models, they are 
also more expensive. In contrast, ENSTIME ensembles, composed of lightweight, 
computationally efficient models, can achieve comparable accuracy at a much lower cost, 
making them an attractive compromise in budget-sensitive applications. Most importantly, 
we find that increasing the number of models in an ensemble does not always yield 
proportional accuracy gains and often results in rapidly escalating costs. Small ensembles 
(with two or three base models) are often sufficient to capture most of the benefit in 
accuracy.

Crucially, our study explored the effects of retraining frequency on ensemble performance 
and cost. Even if these effects are more pronounced for the base models alone, we showed 
that less frequent retraining can reduce computational time, without compromising point 
forecast accuracy, also in the context of ensembles. In many scenarios, 
especially for relatively stable datasets, even the no-retraining setup performed nearly 
as well as continuous retraining, confirming and extending the findings from previous work 
on global models \citep{zanotti1}. Probabilistic forecasting accuracy is more sensitive 
to retraining, especially for higher-frequency datasets like M5, but performance declines 
remain modest for moderate retraining intervals. These insights are highly relevant for 
real-world forecasting systems, where minimizing operational cost without sacrificing 
accuracy is essential. Indeed, organizations can achieve economic benefits by optimizing 
their retraining strategies without compromising the forecast quality.

From a practical standpoint, our findings offer several actionable guidelines for 
forecasters and organizations deploying large-scale forecasting systems. First, ensembles 
should be kept small and strategically designed, prioritizing accuracy or efficiency 
depending on the business context. Second, frequent retraining is often unnecessary.
Overall, monthly retraining emerges as a practical compromise for balancing probabilistic 
forecast accuracy with computational costs. However, if the primary forecasting objective 
is point prediction, even less frequent retraining intervals can be adopted without 
significantly affecting performance. Third, ENSTIME strategies represent a viable, 
low-cost pathway to ensemble forecasting, especially when forecast robustness is desired 
without substantial computational investment. Retailers and other large-scale forecasters 
can leverage these insights to build forecasting systems that are both accurate and 
sustainable. Indeed, these findings also carry broader implications for the sustainability 
of cross-learning driven forecasting systems. Reducing the frequency of retraining not 
only lowers operational costs but also leads to significant energy savings, thereby 
enhancing the environmental sustainability of forecasting processes. This aligns with the 
principles of "Green AI," which advocates for the responsible use of computational 
resources to minimize the ecological footprint of machine learning applications.

While our findings provide strong evidence on the effects of ensembling global forecasting 
models, some limitations remain. First, we restricted our experiments to two retail datasets, 
which, while comprehensive and realistic, may not capture the full diversity of time series 
behaviors found in other domains such as finance, health care, or energy. 
Additionally, this study operates under the assumption that the data-generating process remains 
stable, without notable trends or concept drift. However, in many real-world scenarios, this 
assumption may not hold, potentially affecting the reliability of less frequent retraining 
strategies. Moreover, our ensemble strategies used simple averaging without dynamic weighting 
or more sophisticated stacking techniques, which may further improve performance at the cost 
of added complexity and computation time. We also assumed static cost estimates for computational 
resources, whereas real-world costs may vary based on cloud infrastructure, parallelism, or 
vendor-specific pricing. 
We also deliberately focused on a global modeling approach only, but local or hybrid frameworks 
may offer different cost-accuracy trade-offs worth exploring.
Finally, we encourage further studies to explore the possible trade-off between cost and 
forecast stability in the context of ensemble learning. In fact, the stability of forecasts is 
a highly relevant yet underexplored topic, with important implications for model selection in 
time series forecasting.

In conclusion, our study reinforces the value of ensemble forecasting in global modeling, 
showing that well-designed ensemble strategies, particularly those leveraging periodic 
retraining and lightweight components, can deliver high accuracy at a fraction of the 
computational cost. Striking the right balance between accuracy, efficiency, and retraining 
frequency is key to deploying scalable and sustainable forecasting systems in practice.





\bibliography{refs}


\newpage
\section*{Supplementary material}

In this section, we provide supplementary details on the design of our experiment,
together with tables and figures related to the empirical results of the M5 and VN1 
datasets.


\subsection*{Conformal inference}

The Conformal Inference framework introduced by \citet{vovk2005} offers 
a flexible tool for uncertainty quantification, and has been successfully applied to 
time series forecasting problems \citep{conformalts}.
Since most of the machine learning and deep learning models we use do not natively produce 
probabilistic outputs, we employed Conformal Inference to generate prediction 
intervals. Conformal Inference is a versatile framework that quantifies 
predictive uncertainty using a calibration (validation) set, without relying 
on strong distributional assumptions \citep{vovk2005}. Originally developed 
for i.i.d. data, it has recently been adapted also to time series contexts 
\citep{conformalts}. Its properties, distribution-free, model-agnostic, 
efficient, and suitable for small datasets, make it especially well-suited 
for forecasting benchmarks involving diverse models and datasets.
An alternative approach is to train models using quantile loss functions to 
directly produce quantile forecasts. While this method can effectively capture 
uncertainty, as demonstrated in the M5 Uncertainty competition \citep{m5unc, m5uncsol1},
it requires training a separate model for each quantile level. This significantly 
increases the computational burden, making it less suitable for large-scale 
forecasting applications where efficiency and sustainability are critical considerations.
Moreover, although multi-quantile loss functions can generate multiple quantiles and 
the corresponding prediction intervals, the resulting intervals are often uncalibrated, 
that is, the proportion of observations falling within them may not match the intended 
confidence level. 
For these multiple reasons, we chose to use Conformal Inference as a general framework 
for producing quantile forecasts.
To ensure valid estimation of predictive uncertainty, conformal prediction intervals
were computed on validation sets at least twice the length of the forecast horizon. 
This requirement limited the number of time series we could retain in the dataset 
as explained in previous sections
We used a validation set four times larger than the forecast horizon for
the M5 dataset and double the forecast horizon for the VN1 dataset. 
This limited the number of series from 30,490 to 28,298 for the M5 dataset.

\subsection*{Empirical results}

Table \ref{tab:models} shows the ten different global forecasting models we trained 
in our experiments. For the cost of computation, the Random Forest and the LSTM models 
have been tested on the VN1 weekly dataset only.
Table \ref{tab:ensemble} shows the composition of the eight different 
ensembles for each dataset. 

The Tables \ref{tab:m5_rmsse_ens} and \ref{tab:m5_smql_ens}
show the forecast accuracy of the different ensembles along the 
examined retrain scenarios for the M5 daily dataset, while Table
\ref{tab:m5_ct_ens} depicts the computing time in seconds.

The Tables \ref{tab:vn1_rmsse_ens} and \ref{tab:vn1_smql_ens}
show the forecast accuracy of the different ensembles along the 
examined retrain scenarios for the VN1 weekly dataset, while Table
\ref{tab:vn1_ct_ens} depicts the computing time in seconds.

\begin{figure}
    \centering
    \includegraphics[scale=0.8]{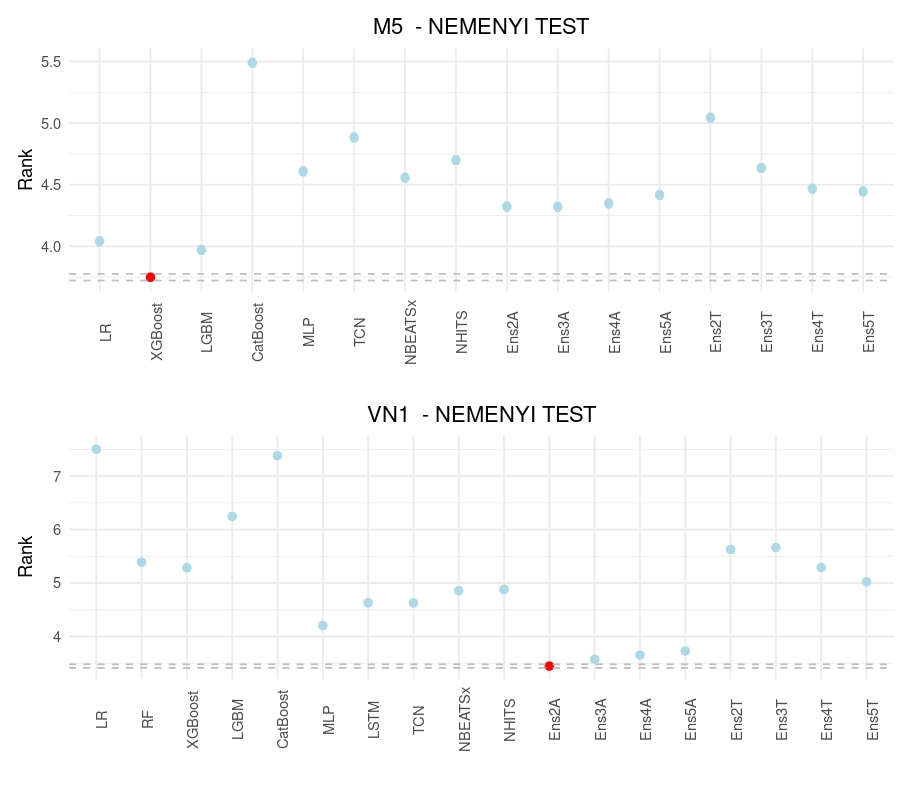}
    \caption{Friedman-Nemenyi test results based on RMSSE.
    The baseline scenarios are \(r = 7\) for the 
    M5 dataset and \(r = 1\) for the VN1 dataset.}
    \label{fig:rmsse_test}
\end{figure}

\begin{figure}
    \centering
    \includegraphics[scale=0.8]{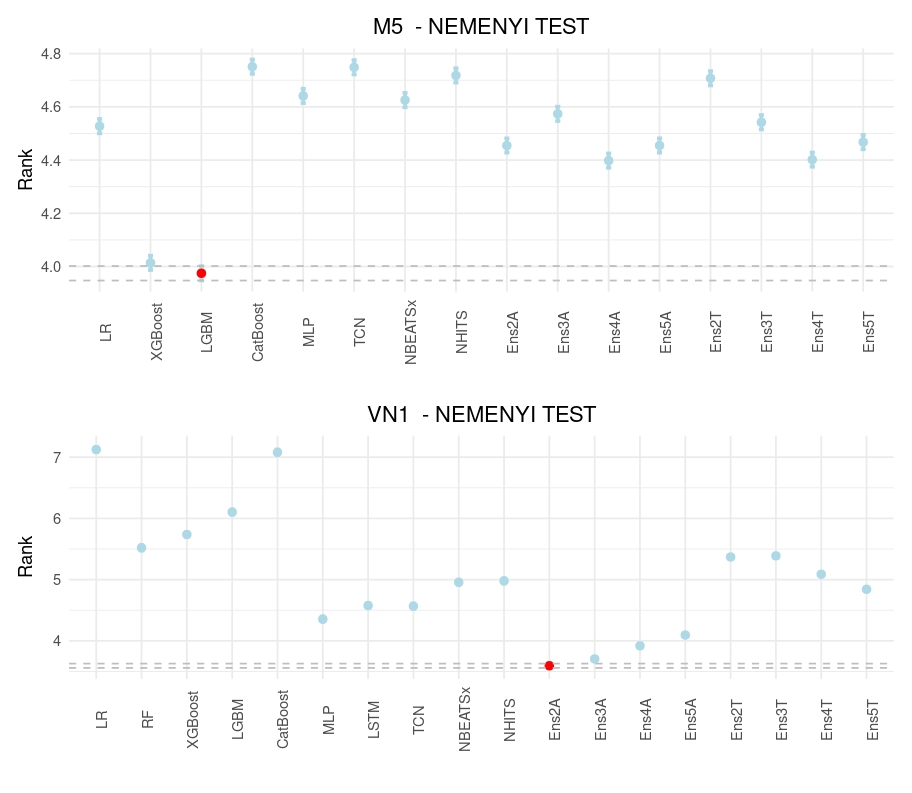}
    \caption{Friedman-Nemenyi test results based on SMQL.
    The baseline scenarios are \(r = 7\) for the 
    M5 dataset and \(r = 1\) for the VN1 dataset.}
    \label{fig:smql_test}
\end{figure}

Figures \ref{fig:rmsse_test} and \ref{fig:smql_test} show the results of the 
Friedman-Nemenyi test for both point forecast and probabilistic accuracy for 
the M5 and VN1 datasets, with respect to the baseline scenario. 

\begin{figure}
    \centering
    \includegraphics[scale=0.55]{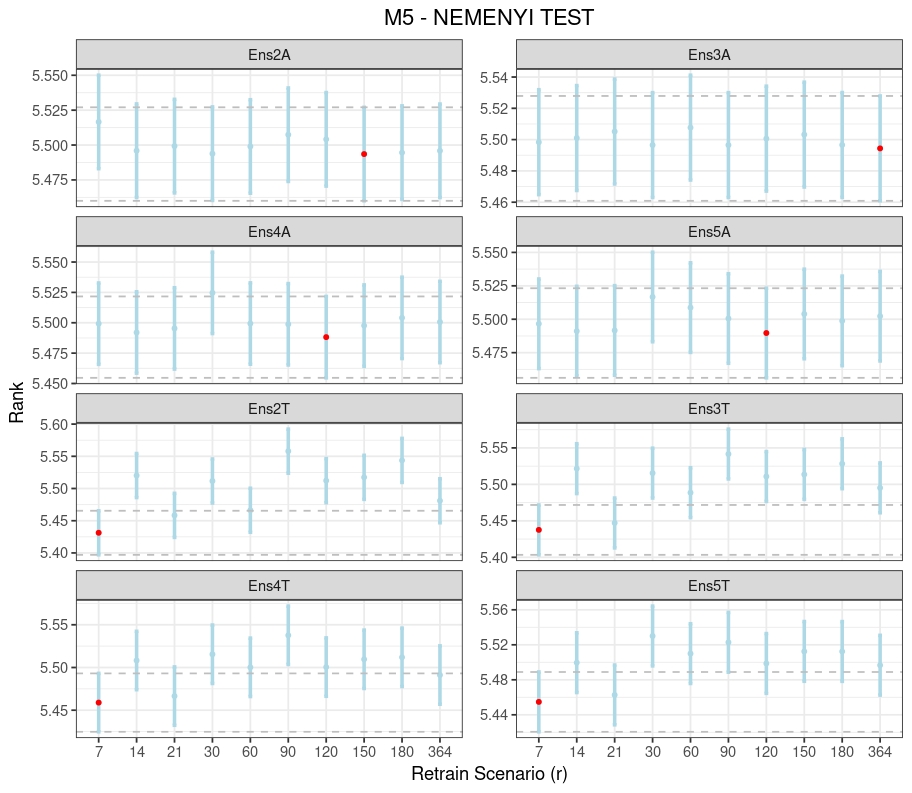}
    \caption{M5 Friedman-Nemenyi test results based on RMSSE.}
    \label{fig:m5_rmsse_test}
\end{figure}

\begin{figure}
    \centering
    \includegraphics[scale=0.55]{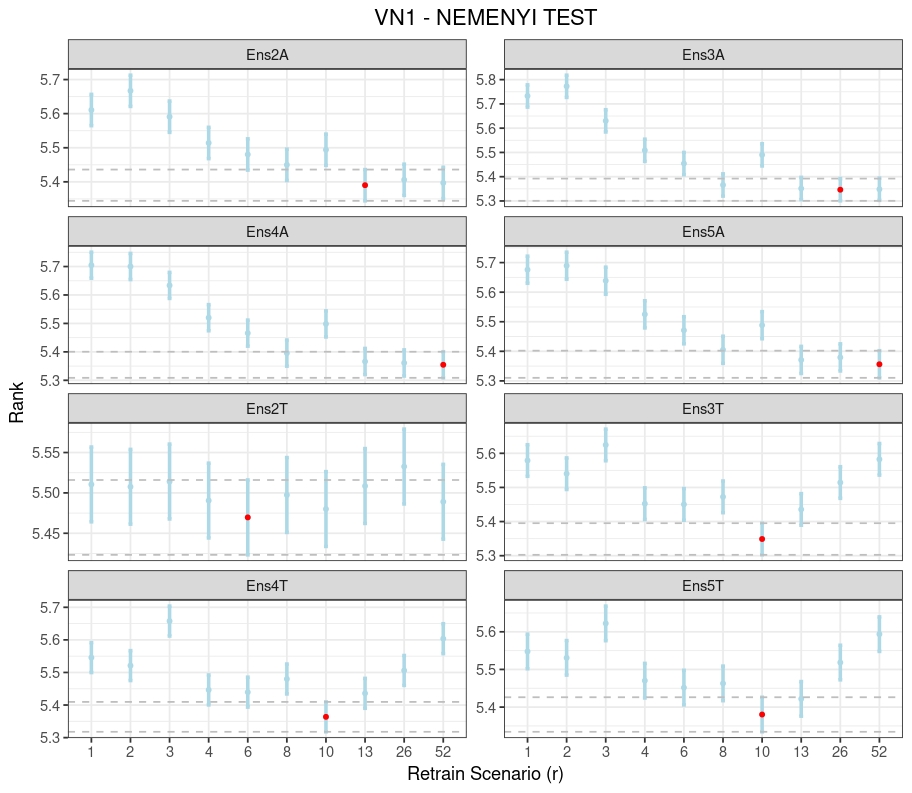}
    \caption{VN1 Friedman-Nemenyi test results based on RMSSE.}
    \label{fig:vn1_rmsse_test}
\end{figure}

\begin{figure}
    \centering
    \includegraphics[scale=0.55]{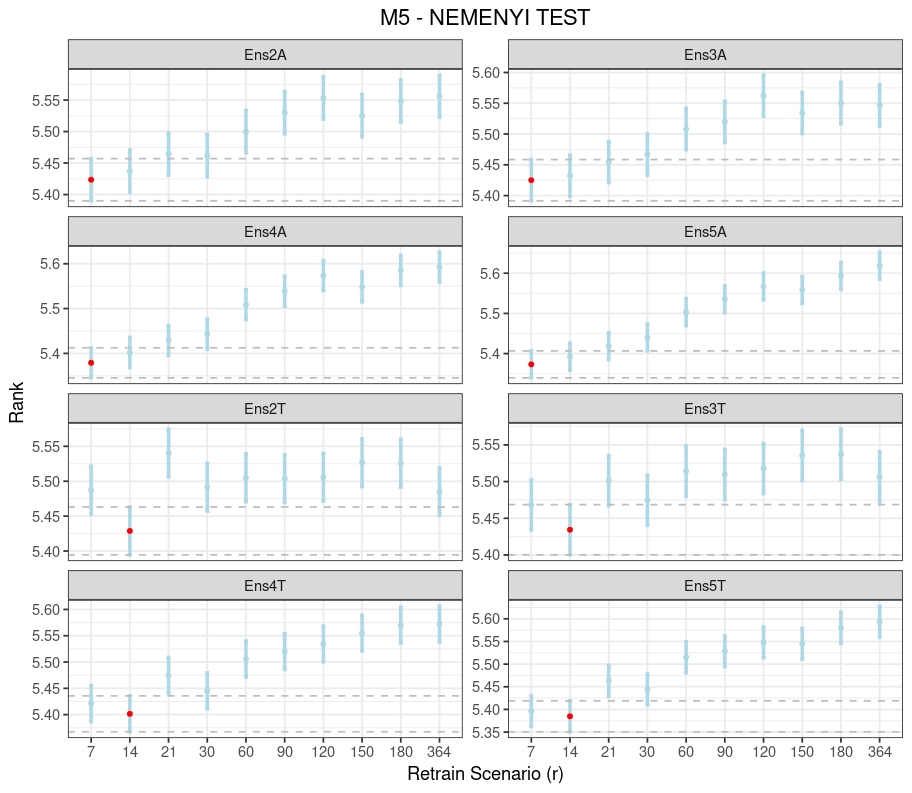}
    \caption{M5 Friedman-Nemenyi test results based on SMQL.}
    \label{fig:m5_smql_test}
\end{figure}

\begin{figure}
    \centering
    \includegraphics[scale=0.55]{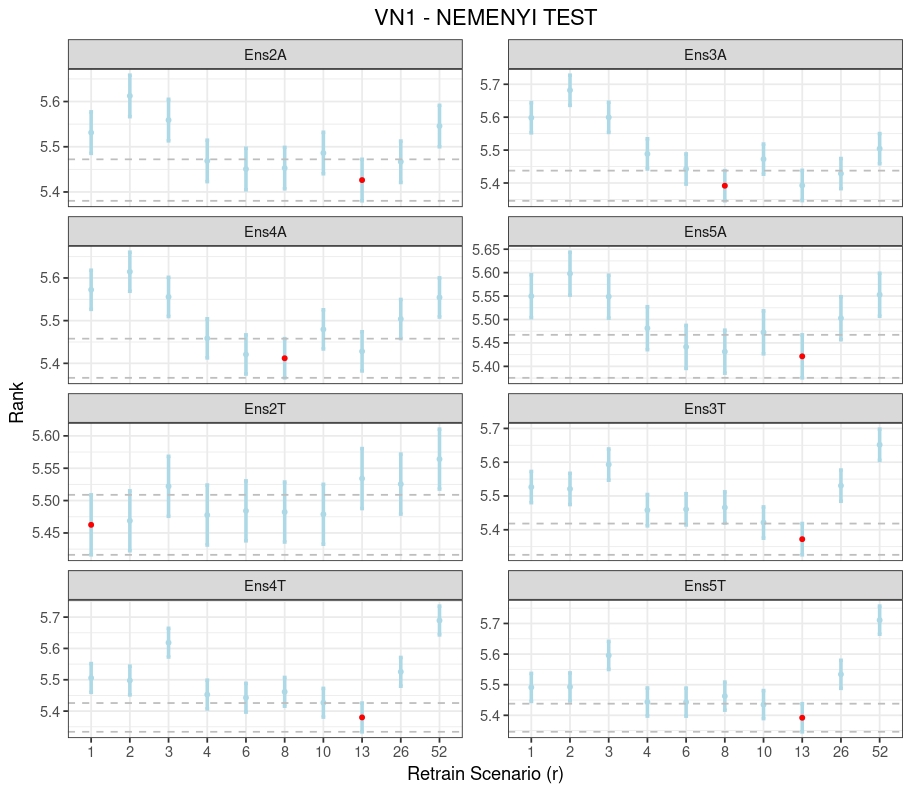}
    \caption{VN1 Friedman-Nemenyi test results based on SMQL.}
    \label{fig:vn1_smql_test}
\end{figure}

Figures \ref{fig:m5_rmsse_test} and \ref{fig:vn1_rmsse_test} show the results of the 
Friedman-Nemenyi test in the context of point forecast accuracy for the M5 and VN1
datasets respectively. Figures \ref{fig:m5_smql_test} and \ref{fig:vn1_smql_test} 
present the test results related to the probabilistic evaluation. 

\begin{figure}
    \centering
    \includegraphics[scale=0.4]{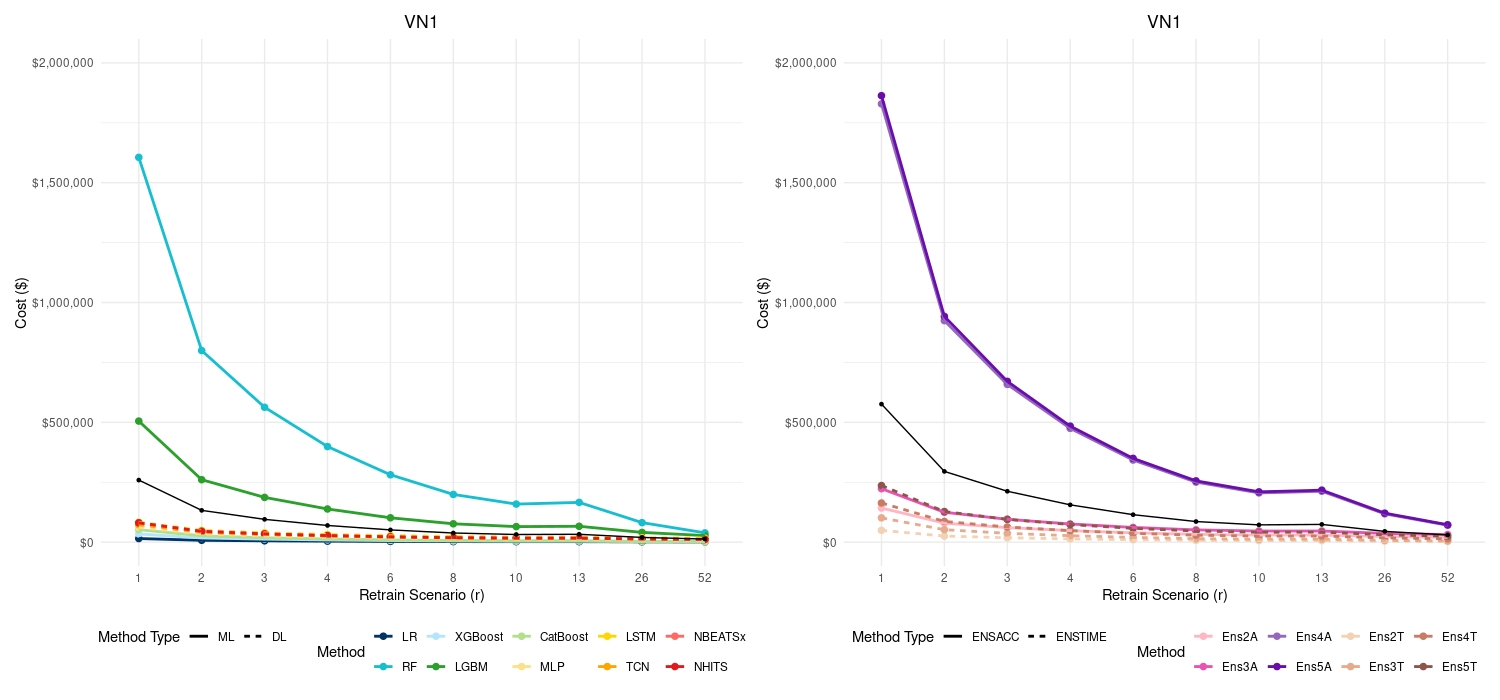}
    \caption{Cost analysis. Comparison between base models and ensembles
    for the VN1 dataset. 
    Each method and retrain scenario combination are shown 
    in relative terms with respect to the baseline scenario, 
    \(r = 1\).}
    \label{fig:vn1_cost_plot}
\end{figure}

The costs tables \ref{tab:m5_costs_base} and \ref{tab:m5_costs_ens}
show the estimated cost in real values of each scenario for 
M5 daily data. 
The costs tables \ref{tab:vn1_costs_base} and \ref{tab:vn1_costs_ens}
show the estimated cost in real values of each scenario for the VN1 weekly data. 
Figure \ref{fig:vn1_cost_plot} shows the cost profiles of each model across
the retraining scenarios for the VN1 dataset.

\begin{table}[ht]
    \centering
    \caption{Global forecasting models used as base models in the experiment.}
    \begin{tabular}{ll}
    \toprule
        \textbf{Machine Learning} & \textbf{Deep Learning} \\
    \midrule
        Linear Regression (LR) \citep{monashrepo} & MLP \citep{mlp} \\
        Random Forest (RF) \citep{randomforest} & LSTM \citep{lstm} \\
        XGBoost \citep{xgb} & TCN \citep{tcn} \\
        LGBM \citep{lgbm} & NBEATS \citep{nbeats} \\
        CatBoost \citep{catboost} & NHITS \citep{nhits} \\
    \bottomrule
    \end{tabular}
    \label{tab:models}
\end{table}

\begin{table}
    \centering
    \caption{Composition of ENSACC and ENSTIME ensembles for the M5 and VN1 datasets.}
    \label{tab:ensemble}
    \begin{tabular}{lll}
    \toprule
    \textbf{Dataset} & \textbf{Ensemble} & \textbf{Models Included} \\
    \midrule
    
    \multirow{10}{*}{M5} 
    & \multicolumn{2}{l}{\textbf{ENSACC (Accuracy-driven combinations)}} \\
    & ENS2A & XGBoost, LGBM \\
    & ENS3A & XGBoost, LGBM, LR \\
    & ENS4A & XGBoost, LGBM, LR, NBEATSx \\
    & ENS5A & XGBoost, LGBM, LR, NBEATSx, MLP \\[0.2cm]
    & \multicolumn{2}{l}{\textbf{ENSTIME (Time-efficient combinations)}} \\
    & ENS2T & CatBoost, LR \\
    & ENS3T & CatBoost, LR, XGBoost \\
    & ENS4T & CatBoost, LR, XGBoost, MLP \\
    & ENS5T & CatBoost, LR, XGBoost, MLP, NBEATSx \\
    
    \midrule
    
    \multirow{10}{*}{VN1} 
    & \multicolumn{2}{l}{\textbf{ENSACC (Accuracy-driven combinations)}} \\
    & ENS2A & MLP, NBEATSx \\
    & ENS3A & MLP, NBEATSx, NHITS \\
    & ENS4A & MLP, NBEATSx, NHITS, RF \\
    & ENS5A & MLP, NBEATSx, NHITS, RF, XGBoost \\[0.2cm]
    & \multicolumn{2}{l}{\textbf{ENSTIME (Time-efficient combinations)}} \\
    & ENS2T & LR, XGBoost \\
    & ENS3T & LR, XGBoost, CatBoost \\
    & ENS4T & LR, XGBoost, CatBoost, MLP \\
    & ENS5T & LR, XGBoost, CatBoost, MLP, TCN \\
    
    \bottomrule
    \end{tabular}
\end{table}

\begin{table}
    \centering
    \begin{tabular}{lrrrrrrrrrr}
      \hline
    Method & 7 & 14 & 21 & 30 & 60 & 90 & 120 & 150 & 180 & 364 \\ 
      \hline
Ens2A & 0.741 & 0.741 & 0.741 & 0.741 & 0.741 & 0.740 & 0.740 & 0.740 & 0.740 & 0.740 \\ 
  Ens3A & 0.741 & 0.741 & 0.741 & 0.741 & 0.741 & 0.740 & 0.741 & 0.741 & 0.740 & 0.740 \\ 
  Ens4A & 0.743 & 0.742 & 0.742 & 0.743 & 0.742 & 0.742 & 0.742 & 0.742 & 0.742 & 0.742 \\ 
  Ens5A & 0.748 & 0.747 & 0.747 & 0.748 & 0.747 & 0.747 & 0.747 & 0.747 & 0.747 & 0.747 \\ 
  Ens2T & 0.798 & 0.810 & 0.799 & 0.809 & 0.805 & 0.812 & 0.809 & 0.809 & 0.810 & 0.806 \\ 
  Ens3T & 0.762 & 0.770 & 0.762 & 0.769 & 0.768 & 0.771 & 0.769 & 0.769 & 0.770 & 0.767 \\ 
  Ens4T & 0.751 & 0.756 & 0.751 & 0.756 & 0.756 & 0.757 & 0.755 & 0.756 & 0.756 & 0.755 \\ 
  Ens5T & 0.750 & 0.754 & 0.749 & 0.756 & 0.754 & 0.755 & 0.753 & 0.754 & 0.754 & 0.753 \\ 
       \hline
    \end{tabular}
    \caption{M5 RMSSE values for each method and retrain scenario combination.}
    \label{tab:m5_rmsse_ens}
\end{table}

\begin{table}
    \centering
    \begin{tabular}{lrrrrrrrrrr}
      \hline
    Method & 7 & 14 & 21 & 30 & 60 & 90 & 120 & 150 & 180 & 364 \\ 
      \hline
Ens2A & 0.244 & 0.244 & 0.245 & 0.245 & 0.246 & 0.247 & 0.248 & 0.247 & 0.248 & 0.249 \\ 
  Ens3A & 0.248 & 0.249 & 0.249 & 0.249 & 0.251 & 0.251 & 0.253 & 0.252 & 0.253 & 0.254 \\ 
  Ens4A & 0.241 & 0.242 & 0.242 & 0.243 & 0.245 & 0.246 & 0.247 & 0.246 & 0.248 & 0.250 \\ 
  Ens5A & 0.242 & 0.243 & 0.243 & 0.244 & 0.246 & 0.247 & 0.248 & 0.248 & 0.249 & 0.251 \\ 
  Ens2T & 0.257 & 0.255 & 0.258 & 0.256 & 0.257 & 0.257 & 0.257 & 0.258 & 0.258 & 0.258 \\ 
  Ens3T & 0.249 & 0.247 & 0.250 & 0.248 & 0.250 & 0.250 & 0.251 & 0.251 & 0.251 & 0.252 \\ 
  Ens4T & 0.242 & 0.241 & 0.243 & 0.243 & 0.245 & 0.245 & 0.246 & 0.246 & 0.247 & 0.248 \\ 
  Ens5T & 0.243 & 0.242 & 0.245 & 0.244 & 0.246 & 0.247 & 0.247 & 0.248 & 0.249 & 0.251 \\ 
       \hline
    \end{tabular}
    \caption{M5 SMQL values for each method and retrain scenario combination.}
    \label{tab:m5_smql_ens}
\end{table}

\begin{table}
    \centering
    \begin{tabular}{lrrrrrrrrrr}
      \hline
    Method & 7 & 14 & 21 & 30 & 60 & 90 & 120 & 150 & 180 & 364 \\ 
      \hline
Ens2A & 59846 & 32015 & 23124 & 17659 & 10899 & 8679 & 7551 & 7717 & 6374 & 5417 \\ 
  Ens3A & 71219 & 38084 & 27513 & 20935 & 12911 & 10197 & 8962 & 9076 & 7507 & 6382 \\ 
  Ens4A & 92445 & 54416 & 42478 & 34519 & 25299 & 22166 & 20632 & 20758 & 18966 & 17648 \\ 
  Ens5A & 110029 & 68460 & 55343 & 46738 & 36710 & 33322 & 31713 & 31730 & 29678 & 28287 \\ 
  Ens2T & 21797 & 11591 & 8297 & 6178 & 3705 & 2803 & 2494 & 2472 & 2025 & 1674 \\ 
  Ens3T & 37214 & 19780 & 14141 & 10591 & 6272 & 4826 & 4243 & 4223 & 3427 & 2834 \\ 
  Ens4T & 54797 & 33824 & 27005 & 22811 & 17684 & 15982 & 15323 & 15196 & 14139 & 13473 \\ 
  Ens5T & 76024 & 50156 & 41971 & 36394 & 30071 & 27951 & 26994 & 26878 & 25598 & 24739 \\ 
       \hline
    \end{tabular}
    \caption{M5 CT values for each method and retrain scenario combination.}
    \label{tab:m5_ct_ens}
\end{table}

\begin{table}
    \centering
    \begin{tabular}{lrrrrrrrrrr}
      \hline
    Method & 1 & 2 & 3 & 4 & 6 & 8 & 10 & 13 & 26 & 52 \\ 
      \hline
Ens2A & 1.478 & 1.485 & 1.486 & 1.459 & 1.450 & 1.450 & 1.453 & 1.437 & 1.434 & 1.437 \\ 
  Ens3A & 1.523 & 1.526 & 1.510 & 1.468 & 1.458 & 1.441 & 1.456 & 1.435 & 1.435 & 1.437 \\ 
  Ens4A & 1.530 & 1.533 & 1.520 & 1.478 & 1.467 & 1.451 & 1.465 & 1.442 & 1.441 & 1.443 \\ 
  Ens5A & 1.550 & 1.549 & 1.546 & 1.501 & 1.490 & 1.474 & 1.485 & 1.465 & 1.465 & 1.462 \\ 
  Ens2T & 5.694 & 5.655 & 5.647 & 5.563 & 5.513 & 5.618 & 5.578 & 5.589 & 5.680 & 5.257 \\ 
  Ens3T & 5.286 & 5.297 & 5.262 & 5.263 & 5.290 & 5.253 & 4.943 & 5.118 & 5.454 & 5.503 \\ 
  Ens4T & 4.447 & 4.406 & 4.620 & 4.276 & 4.236 & 4.287 & 4.144 & 4.151 & 4.391 & 4.635 \\ 
  Ens5T & 3.664 & 3.609 & 3.792 & 3.506 & 3.478 & 3.519 & 3.407 & 3.403 & 3.600 & 3.781 \\
       \hline
    \end{tabular}
    \caption{VN1 RMSSE values for each method and retrain scenario combination.}
    \label{tab:vn1_rmsse_ens}
\end{table}

\begin{table}
    \centering
    \begin{tabular}{lrrrrrrrrrr}
      \hline
    Method & 1 & 2 & 3 & 4 & 6 & 8 & 10 & 13 & 26 & 52 \\ 
      \hline
Ens2A & 0.826 & 0.847 & 0.840 & 0.815 & 0.807 & 0.808 & 0.819 & 0.801 & 0.814 & 0.844 \\ 
  Ens3A & 0.867 & 0.885 & 0.863 & 0.829 & 0.818 & 0.802 & 0.828 & 0.803 & 0.816 & 0.845 \\ 
  Ens4A & 0.926 & 0.941 & 0.927 & 0.895 & 0.888 & 0.879 & 0.900 & 0.885 & 0.903 & 0.931 \\ 
  Ens5A & 1.013 & 1.024 & 1.014 & 0.990 & 0.982 & 0.975 & 0.989 & 0.971 & 0.999 & 1.018 \\ 
  Ens2T & 4.388 & 4.392 & 4.443 & 4.370 & 4.379 & 4.353 & 4.325 & 4.419 & 4.377 & 4.110 \\ 
  Ens3T & 3.972 & 4.002 & 4.026 & 4.030 & 4.007 & 3.905 & 3.777 & 3.649 & 4.003 & 4.270 \\ 
  Ens4T & 3.420 & 3.403 & 3.668 & 3.321 & 3.248 & 3.254 & 3.226 & 3.038 & 3.284 & 3.771 \\ 
  Ens5T & 2.832 & 2.805 & 3.011 & 2.738 & 2.687 & 2.691 & 2.673 & 2.513 & 2.725 & 3.105 \\ 
       \hline
    \end{tabular}
    \caption{VN1 SMQL values for each method and retrain scenario combination.}
    \label{tab:vn1_smql_ens}
\end{table}

\begin{table}
    \centering
    \begin{tabular}{lrrrrrrrrrr}
      \hline
    Method & 1 & 2 & 3 & 4 & 6 & 8 & 10 & 13 & 26 & 52 \\ 
      \hline
    Ens2A & 2205 & 1228 & 940 & 748 & 607 & 508 & 460 & 460 & 361 & 312 \\ 
      Ens3A & 3456 & 1932 & 1477 & 1175 & 952 & 796 & 722 & 721 & 569 & 492 \\ 
      Ens4A & 28318 & 14312 & 10192 & 7352 & 5307 & 3880 & 3183 & 3291 & 1831 & 1094 \\ 
      Ens5A & 28849 & 14586 & 10391 & 7498 & 5414 & 3962 & 3252 & 3361 & 1876 & 1125 \\ 
      Ens2T & 766 & 395 & 287 & 212 & 157 & 120 & 102 & 104 & 67 & 48 \\ 
      Ens3T & 1572 & 807 & 574 & 419 & 306 & 229 & 192 & 197 & 118 & 79 \\ 
      Ens4T & 2533 & 1341 & 983 & 747 & 571 & 452 & 394 & 398 & 277 & 217 \\ 
      Ens5T & 3660 & 1980 & 1477 & 1141 & 893 & 724 & 641 & 652 & 477 & 392 \\ 
       \hline
    \end{tabular}
    \caption{VN1 CT values for each method and retrain scenario combination.}
    \label{tab:vn1_ct_ens}
\end{table}

\begin{table}[ht]
    \centering
    \begin{adjustbox}{max width=\textwidth}
    \begin{tabular}{lcccccccccc}
    \toprule
    Method & 7 & 14 & 21 & 30 & 60 & 90 & 120 & 150 & 180 & 364 \\
    \midrule
    LR & 390,732 & 208,499 & 150,791 & 112,547 & 69,132 & 52,131 & 48,474	& 46,678 & 38,937 &	33,151 \\
    XGBoost & 529,679 & 281,358 & 200,778 & 151,635 & 88,200 & 69,495 & 60,082 & 60,155 & 48,189 & 39,861 \\
    LGBM & 1,526,424 & 818,569 & 593,676 & 455,075 & 286,262 & 228,698 & 199,337 & 204,972 & 170,802 & 146,252 \\
    CatBoost & 358,123 & 189,714 & 134,258 & 99,699 & 58,167 & 44,164 & 37,206 & 38,266 & 30,622 & 24,362 \\
    MLP & 604,120 & 482,505 & 441,981 & 419,823 & 392,046 & 383,287 & 380,697 & 376,968 & 368,023 & 365,525 \\
    TCN & 1,146,256 & 960,626 & 890,924 & 862,432 & 815,006 & 801,612 & 791,801 & 791,348 & 784,201 & 776,181 \\
    NBEATSx & 729,263 & 561,105 & 514,166 & 466,683 & 425,593 & 411,232 & 400,953 & 401,352 & 393,673 & 387,058 \\
    NHITS & 754,783 & 573,233 & 510,655 & 470,077 & 425,684 & 409,844 & 402,731 & 402,394 & 393,071 & 385,470 \\
    Average & 754,922 & 509,451 & 429,654 & 379,746 & 320,011 & 300,058 & 290,160 & 290,267 & 278,440 & 269,733 \\
    \bottomrule
    \end{tabular}
    \end{adjustbox}
    \caption{M5 estimated costs for each base method and retrain scenario combination (in \$).}
    \label{tab:m5_costs_base}
\end{table}

\begin{table}
    \centering
    \begin{adjustbox}{max width=\textwidth}
    \begin{tabular}{lcccccccccc}
    \toprule
    Method & 7 & 14 & 21 & 30 & 60 & 90 & 120 & 150 & 180 & 364 \\
    \midrule
    Ens2A & 2,056,103 & 1,099,927 & 794,454 & 606,710 & 374,461 & 298,193 & 259,419 & 265,127 & 218,991 & 186,113 \\ 
  Ens3A & 2,446,835 & 1,308,426 & 945,244 & 719,256 & 443,594 & 350,323 & 307,893 & 311,805 & 257,928 & 219,264 \\ 
  Ens4A & 3,176,097 & 1,869,531 & 1,459,411 & 1,185,940 & 869,187 & 761,555 & 708,845 & 713,157 & 651,601 & 606,322 \\ 
  Ens5A & 3,780,218 & 2,352,036 & 1,901,392 & 1,605,763 & 1,261,233 & 1,144,843 & 1,089,543 & 1,090,125 & 1,019,624 & 971,847 \\ 
  Ens2T & 748,854 & 398,213 & 285,049 & 212,246 & 127,299 & 96,295 & 85,680 & 84,944 & 69,559 & 57,513 \\ 
  Ens3T & 1,278,533 & 679,571 & 485,827 & 363,881 & 215,499 & 165,789 & 145,762 & 145,099 & 117,748 & 97,375 \\ 
  Ens4T & 1,882,653 & 1,162,076 & 927,808 & 783,704 & 607,545 & 549,077 & 526,459 & 522,067 & 485,770 & 462,900 \\ 
  Ens5T & 2,611,916 & 1,723,181 & 1,441,975 & 1,250,387 & 1,033,138 & 960,309 & 927,412 & 923,419 & 879,444 & 849,958 \\
  Average & 2,247,651 & 1,324,120 & 1,030,144 & 840,985 & 616,494 & 540,797 & 506,376 & 506,967 & 462,583 & 431,411 \\
    \bottomrule
    \end{tabular}
    \end{adjustbox}
    \caption{M5 estimated costs for each ensemble method and retrain scenario combination (in \$).}
    \label{tab:m5_costs_ens}
\end{table}

\begin{table}
    \centering
    \begin{adjustbox}{max width=\textwidth}
    \begin{tabular}{lcccccccccc}
        \toprule
        Method & 1 & 2 & 3 & 4 & 6 & 8 & 10 & 13 & 26 & 52 \\
        \midrule
        LR & 15,234 & 7,839 & 5,731 & 4,292 & 3,205 & 2,508 & 2,140 & 2,190 & 1,463 & 1,099 \\
        RF & 1,605,768 & 799,597 & 562,896 & 398,954 & 281,246 & 199,214 & 158,959 & 165,975 & 81,529 & 38,865 \\
        XGBoost & 34,254 & 17,687 & 12,796 & 9,413 & 6,944 & 5,262 & 4,473 & 4,534 & 2,863 & 2,005 \\
        LGBM & 505,304 & 260,721 & 186,904 & 138,460 & 101,775 & 76,738 & 65,013 & 66,335 & 40,721 & 27,701 \\
        CatBoost & 52,021 & 26,594 & 18,515 & 13,366 & 9,611 & 7,022 & 5,798 & 5,977 & 3,311 & 2,023 \\
        MLP & 62,102 & 34,489 & 26,431 & 21,149 & 17,121 & 14,391 & 13,027 & 13,015 & 10,255 & 8,891 \\
        LSTM & 82,927 & 49,019 & 38,851 & 31,962 & 27,022 & 23,465 & 21,829 & 21,932 & 18,415 & 16,652 \\
        TCN & 72,763 & 41,291 & 31,910 & 25,464 & 20,791 & 17,579 & 15,983 & 16,365 & 12,936 & 11,284 \\
        NBEATSx & 80,337 & 44,837 & 34,283 & 27,179 & 22,057 & 18,387 & 16,686 & 16,702 & 13,085 & 11,277 \\
        NHITS & 80,776 & 45,458 & 34,688 & 27,585 & 22,336 & 18,607 & 16,888 & 16,855 & 13,405 & 11,601 \\
        Average & 259,149 & 132,753 & 95,301 & 69,782 & 51,211 & 38,317 & 32,080 & 32,988 & 19,798 & 13,140 \\
        \bottomrule
    \end{tabular}
    \end{adjustbox}
    \caption{VN1 estimated costs for each base method and retrain scenario combination (in \$).}
    \label{tab:vn1_costs_base}
\end{table}

\begin{table}
    \centering
    \begin{adjustbox}{max width=\textwidth}
    \begin{tabular}{lcccccccccc}
        \toprule
        Method & 1 & 2 & 3 & 4 & 6 & 8 & 10 & 13 & 26 & 52 \\
        \midrule
Ens2A & 142,439 & 79,326 & 60,714 & 48,328 & 39,177 & 32,778 & 29,713 & 29,717 & 23,340 & 20,167 \\ 
  Ens3A & 223,214 & 124,784 & 95,402 & 75,913 & 61,513 & 51,385 & 46,602 & 46,572 & 36,745 & 31,769 \\ 
  Ens4A & 1,828,982 & 924,381 & 658,299 & 474,867 & 342,760 & 250,599 & 205,561 & 212,547 & 118,274 & 70,634 \\ 
  Ens5A & 1,863,237 & 942,068 & 671,095 & 484,280 & 349,703 & 255,861 & 210,033 & 217,080 & 121,137 & 72,638 \\ 
  Ens2T & 49,488 & 25,526 & 18,527 & 13,705 & 10,149 & 7,769 & 6,613 & 6,724 & 4,327 & 3,103 \\ 
  Ens3T & 101,509 & 52,120 & 37,042 & 27,071 & 19,760 & 14,792 & 12,410 & 12,700 & 7,637 & 5,126 \\ 
  Ens4T & 163,611 & 86,609 & 63,473 & 48,220 & 36,880 & 29,183 & 25,437 & 25,715 & 17,892 & 14,017 \\ 
  Ens5T & 236,374 & 127,900 & 95,383 & 73,684 & 57,671 & 46,762 & 41,420 & 42,080 & 30,828 & 25,301 \\ 
  Average & 576,106 & 295,339 & 212,491 & 155,758 & 114,701 & 86,140 & 72,223 & 74,141 & 45,022 & 30,344 \\
        \bottomrule
    \end{tabular}
    \end{adjustbox}
    \caption{VN1 estimated costs for each ensemble method and retrain scenario combination (in \$).}
    \label{tab:vn1_costs_ens}
\end{table}

\end{document}